\documentclass{IEEEtran}
\UseRawInputEncoding

\usepackage{enumitem,kantlipsum}
\usepackage{graphicx}
\usepackage{epstopdf}
\usepackage{amssymb}
\usepackage{amsmath}
\usepackage{subfigure}
\usepackage{epsfig}
\usepackage{color}
\usepackage{enumitem}
\usepackage{float}
\usepackage{soul}
\usepackage{wrapfig}
\usepackage{arydshln}

\usepackage{amsthm}

\def\0{{\bf 0}}
\def\1{{\bf 1}}

\def\beq{\begin{equation*}}
    \def\eeq{\end{equation*}}
\def\bql{\begin{equation}}
    \def\eql{\end{equation}}
\def\bqn{\begin{eqnarray*}}
    \def\eqn{\end{eqnarray*}}
\def\bnl{\begin{eqnarray}}
    \def\enl{\end{eqnarray}}
\def\bma{\begin{bmatrix}}
    \def\ema{\end{bmatrix}}
\def\bmx{\begin{matrix}}
    \def\emx{\end{matrix}}
\def\ben{\begin{enumerate}}
    \def\een{\end{enumerate}}
\def\bit{\begin{itemize}}
    \def\eit{\end{itemize}}
\def\bei{\begin{itemize}}
    \def\eei{\end{itemize}}
\def\bet{\begin{tabular}}
    \def\eet{\end{tabular}}

\newcommand{\ba}{\mathbf{a}}

\newcommand{\R}{\mathbb{R}}

\newcommand{\A}{\mathcal{A}}

\newcommand{\be}{\mathbf{e}}

\newcommand{\bu}{\mathbf{u}}

\newcommand{\bv}{\mathbf{v}}

\usepackage[dvipsnames]{xcolor}

\def\R{\mathbb{R}}

\def\1{{\bf1}}

\def\b{{\beta}}
\def\a{\alpha}

\def\bit{\begin{itemize}}
\def\eit{\end{itemize}}

\def\be{\begin{equation}}
\def\ee{\end{equation}}
\def\ba{\begin{eqnarray}}
\def\ea{\end{eqnarray}}

\def\bes{\begin{equation*}}
\def\ees{\end{equation*}}
\def\bas{\begin{eqnarray*}}
\def\eas{\end{eqnarray*}}

\newtheorem{Remark 1}{Remark}
\newtheorem{Remark 2}[Remark 1]{Remark}
\newtheorem{Remark 3}[Remark 1]{Remark}
\newtheorem{Remark 4}[Remark 1]{Remark}
\newtheorem{Remark 5}[Remark 1]{Remark}
\newtheorem{Remark 6}[Remark 1]{Remark}
\newtheorem{Remark 7}[Remark 1]{Remark}

\newtheorem{Lemma 1}{Lemma}
\newtheorem{Lemma 2}[Lemma 1]{Lemma}
\newtheorem{Lemma 3}[Lemma 1]{Lemma}
\newtheorem{Lemma 4}[Lemma 1]{Lemma}
\newtheorem{Lemma 5}[Lemma 1]{Lemma}
\newtheorem{Lemma 6}[Lemma 1]{Lemma}
\newtheorem{Lemma 7}[Lemma 1]{Lemma}

\newtheorem{Assumption 1}{Assumption}
\newtheorem{Assumption 2}[Assumption 1]{Assumption}
\newtheorem{Assumption 3}[Assumption 1]{Assumption}
\newtheorem{Assumption 4}[Assumption 1]{Assumption}
\newtheorem{Definition 1}{Definition}
\newtheorem{Theorem 1}{Theorem}
\newtheorem{Theorem 2}[Theorem 1]{Theorem}
\newtheorem{Theorem 3}[Theorem 1]{Theorem}
\newtheorem{Theorem 4}[Theorem 1]{Theorem}
\newtheorem{Theorem 5}[Theorem 1]{Theorem}
\newtheorem{Theorem 6}[Theorem 1]{Theorem}
\newtheorem{Theorem 7}[Theorem 1]{Theorem}
\newtheorem{Theorem 8}[Theorem 1]{Theorem}
\newtheorem{Theorem 9}[Theorem 1]{Theorem}
\newtheorem{Theorem 10}[Theorem 1]{Theorem}

\title{\LARGE \bf
 Decentralized Stochastic Optimization with Inherent Privacy Protection}

\author{Yongqiang Wang, H. Vincent Poor 
\vspace{-0.8cm}
\thanks{ The work was supported in part by the National Science Foundation under Grants ECCS-1912702 and CCF-2106293.}
\thanks{Yongqiang Wang is with the Department of Electrical and Computer Engineering, Clemson University, Clemson, SC 29634, USA
{\tt\small{yongqiw}@clemson.edu}
}%
\thanks{H. Vincent Poor is with the Department of Electrical Engineering, Princeton
University, Princeton, NJ 08544, USA {\tt\small poor@princeton.edu}}
}

\begin{document}

\maketitle
\thispagestyle{empty}
\pagestyle{empty}

\begin{abstract}
Decentralized stochastic optimization is the basic building block of
modern collaborative machine learning, distributed estimation and
control, and large-scale sensing. Since involved data  usually
contain sensitive information like user locations, healthcare
records and financial transactions,  privacy protection  has become
an increasingly pressing need in the implementation of decentralized
stochastic optimization algorithms. In this paper, we propose a
decentralized stochastic gradient descent algorithm which is
embedded with inherent privacy protection for every participating
agent against other participating agents and external eavesdroppers.
This proposed algorithm builds in a dynamics based
gradient-obfuscation mechanism to enable privacy protection without
compromising optimization accuracy, which is in significant
difference from differential-privacy based privacy solutions for
decentralized optimization that have to trade optimization accuracy
for privacy. The dynamics based privacy approach is encryption-free,
and hence avoids incurring heavy communication or computation
overhead, which is a common problem with encryption based privacy
solutions for decentralized stochastic optimization. Besides
rigorously characterizing the convergence performance of the
proposed decentralized stochastic gradient descent algorithm under
both convex objective functions and non-convex objective functions,
we also provide rigorous information-theoretic analysis of its
strength of  privacy protection. Simulation results for a
distributed estimation problem as well as numerical experiments for
decentralized   learning on a benchmark machine learning dataset
confirm the effectiveness of the proposed approach.
\end{abstract}

%
\section{Introduction}
Decentralized optimization and learning has been an area of
intensive research due to its wide applications in large-scale
sensing \cite{rabbat2004distributed}, distributed estimation and
control \cite{bullo2009distributed,jakovetic2014fast}, and big data
analytics \cite{daneshmand2015hybrid}. In recent years, in order to
handle the enormous growth in   data volumes as well as to address
practical data imperfections, decentralized stochastic gradient
descent (SGD), in which participating agents use {\it noisy} local
gradients and peer-to-peer communications to cooperatively solve a
network optimization problem, is gaining increased attention. For
example, in modern distributed machine learning on massive datasets,
multiple training agents cooperatively  train a neural network where
each participating agent updates a local copy of the neural network
model using both information received from neighboring agents and
local noisy gradients calculated from one (or a small batch of) data
points available to the agent. Computing gradients from only one (or
a small batch of) available data point yields a noisy estimation of
the exact gradient, but is completely necessary because evaluating
the precise gradient using all available data points usually is
extremely expensive in computation or even simply impractical.
Furthermore, with the advent of Internet of things  which features
the explosive growth of low-cost sensing and communication devices,
the data fed to distributed optimization are usually subject to
measurement noises \cite{xin2020decentralized}. As
 deterministic   optimization approaches will typically falter
when involved data are massive or noisy
\cite{bottou2018optimization}, it is mandatory to investigate
  decentralized stochastic
optimization algorithms.

Centralized SGD algorithms  date back to  the 1950s
\cite{robbins1951stochastic} and decentralized SGD    dates back to
the 1980s \cite{tsitsiklis1986distributed}. To date, plenty of
results on decentralized SGD have been reported, for both the case
with convex objective functions (e.g.,
\cite{ram2010distributed,nedic2016stochastic,jakovetic2018convergence,sayin2017stochastic,pu2020distributed,rabbat2015multi,shamir2014distributed,sirb2018decentralized})
and  the case with non-convex objective functions (e.g.,
\cite{bianchi2012convergence,tatarenko2017non,lian2017can,singh2020sparq,koloskova2019decentralized,george2019distributed}).
In such decentralized SGD algorithms, as participating agents only
share gradient/model updates and never let raw data leave
participants' machines, it was believed that the privacy of
participating agents can be protected by the decentralized computing
architecture. However, recent studies revealed a completely
different picture: not only are properties of the data, such as
membership associations, are inferable from  gradient/model updates
shared in decentralized stochastic optimization
\cite{shokri2015privacy,melis2019exploiting}, but even raw data can
be precisely reversely inferred from shared gradients (pixel-wise
accurate for images and token-wise matching for texts)
\cite{zhu2019deep}. This poses a severe threat to the privacy of
participants in decentralized stochastic optimization, because data
used in optimization often contain sensitive information such as
user positions, healthcare records, or financial transactions.

To protect the privacy of participating agents in distributed
stochastic gradient methods, recently several approaches have been
proposed. One approach resorts to differential privacy
\cite{differential_privacy_dwork} which adds additive noise to
shared gradient/model updates (e.g.,
\cite{chaudhuri2011differentially,bassily2014private,abadi2016deep,wu2017bolt,yu2019differentially,bagdasaryan2019differential,mcmahan2018learning,wang2022tailoring}).
\cite{raginsky2017non} and \cite{dwork2015reusable}  indicate  that additive
noise used for differential privacy can even facilitate convergence
to the global optimum in non-convex optimization. However, recent
systematic studies   in \cite{zhu2019deep} indicate that a small
magnitude of differential-privacy noise cannot thwart strong privacy
attacks and differential privacy based defense can achieve
reasonable privacy protection {\it ``only when the noise variance is
large enough to degrade accuracy \cite{zhu2019deep}."} 
Another
commonly used approach to enable privacy in decentralized SGD is to
employ secure multi-party computation approaches such as homomorphic
encryption \cite{paillier1999public} or garbled circuit
\cite{yao1986generate}, which, however, will incur a {\it runtime
overhead of three to four orders of magnitude}
\cite{hynes2018efficient}. Furthermore, homomorphic encryption  can
only protect against a parameter server \cite{zhu2019deep}. Hardware
based approaches such as trusted hardware enclaves have also been
discussed in the literature
\cite{hynes2018efficient,tramer2018slalom,ohrimenko2016oblivious}.
However, similar to homomorphic encryption based approaches,  these
approaches cannot be directly used   to prevent multiple data
providers from inferring each others' data during stochastic
optimization. Recently, several results were reported to exploit the
structural properties of decentralized optimization to enable
privacy. For example, the authors in
  \cite{yan2012distributed,lou2017privacy} showed that privacy can be enabled by
  adding a {\it constant} uncertain parameter in projection or step
  sizes. The authors of \cite{gade2018privacy} showed that network structure can be leveraged to
  construct spatially correlated ``structured" noise to cover
  gradient information.
   Although these
  approaches can ensure  optimization accuracy,
  their enabled privacy is restricted: projection based privacy depends on
  the size of the projection set -- a large projection set nullifies
  privacy protection whereas a small projection set offers strong
  privacy
  protection but requires {\it a priori} knowledge of the optimal solution; ``structured"
  noise based approaches require  each agent to have a certain number
  of neighbors that do not share information with the adversary.  In
  fact, such a structural constraint is required in most existing
  accuracy-friendly privacy solutions to decentralized optimization.
  For example,
   our results in \cite{zhang2019admm} show that
   even the partially homomorphic encryption based privacy
  approaches  require that the adversary doesn't have access to a target agent's communications with all of its neighbors.

Inspired by our recent finding that the privacy of participating
agents in decentralized consensus computations can be ensured by
manipulating inherent consensus dynamics
\cite{ruan2019secure,wang2019privacy,gao2022algorithm}, in this paper, we propose to
enable intrinsic privacy protection in decentralized stochastic
gradient methods by judiciously manipulating the inherent dynamics
of inter-agent information fusion and gradient-descent operations.
More specifically, we let each participating agent  use time-varying
and heterogeneous stepsize and an additional stochastic mixing
coefficient to obscure its gradients when sharing information with
its neighbors. Since the privacy approach will not increase the size
or the number of shared messages, it will not incur heavy
communication or computation overhead. Furthermore, we rigorously
prove that the proposed privacy approach does not affect the
 convergence to the desired solution, which is in distinct
 difference from differential privacy based privacy approaches. {It is worth noting
 that although the approaches in \cite{xu2015augmented,nedic2017geometrically}
   use  heterogeneous stepsizes, they do not allow the stepsizes to be random and time-varying,
   which, however,
   is  key to achieving the enabled privacy here. Moreover, the algorithms in \cite{xu2015augmented,nedic2017geometrically} have to share two
   variables (an optimization variable and a gradient-tracking variable) in every iteration whereas our approach only needs to share one variable, which is important
   since information sharing can create  significant  communication
overhead and even communication bottlenecks in applications like
modern deep learning where the dimension of variables scales  to
hundreds of millions  \cite{tang2020communication}.}

The main contributions  are as follows: 1) We propose an inherently
privacy-preserving  decentralized SGD algorithm. This algorithm can
protect the privacy of all participating agents even when all
messages shared by an agent are accessible to an adversary, a
scenario which fails existing accuracy-friendly privacy-preserving
approaches for decentralized (stochastic) optimization; 2) The
inherently privacy-preserving approach is efficient in communication
and computation in that it is encryption-free and does not increase
the size or the number of shared messages among participating
agents; 3) We prove that the privacy approach does not affect the
accuracy of decentralized stochastic optimization in both the
convex-objective-function case and the non-convex-objective-function
case, which is in distinct difference from difference-privacy based
privacy solutions which trade accuracy for privacy; 4) We propose a
rigorous information-theoretic privacy analysis approach to evaluate
the strength of privacy inherently embedded in the algorithm; 5) We
use both simulations for a decentralized estimation problem and
numerical experiments for decentralized deep learning on a benchmark
dataset to confirm the effectiveness of the proposed results.

The organization of the paper is as follows. Sec. II gives the
problem formulation. Sec. III presents the inherently
privacy-preserving decentralized SGD algorithm. Sec. IV proves
converge of the proposed algorithm when the aggregate objective function is
convex. Sec. V analyzes the convergence performance of the
inherently privacy-preserving decentralized SGD algorithm when the
objective functions are non-convex. Sec. VI proposes an
information-theoretic approach to rigorously analyze the strength of
enabled privacy. Sec. VII gives simulation results as well as
numerical experiments on a benchmark machine learning dataset to
confirm the obtained  results. Finally Sec. VIII concludes the
paper.

{\bf Notation:} We use   $\mathbb{R}$ to denote the set of real
numbers and $\mathbb{R}^d$ the Euclidean space of dimension $d$.
${\bf 1}$ denotes a column vector with all entries equal to 1. A
vector is viewed as a column vector, unless otherwise stated. For a
vector $x$, $x_i$ denotes its $i$th element. $A^T$ denotes the
transpose of matrix $A$ and $x^Ty$ denotes the scalar product of two
vectors $x$ and $y$. {We use $\langle\cdot,\cdot\rangle$
to denote inner product  and $\|\cdot\|$ to denote the  Euclidean
norm for a vector (induced Euclidean norm for a matrix)}. A square
matrix $A$ is said to be column-stochastic when its elements in
every column add up to one. A matrix $A$ is said to be
doubly-stochastic when both $A$ and $A^T$ are column-stochastic
matrices. We use $P(\mathcal{A})$ to denote the probability of an
event $\mathcal{A}$ and $\mathbb{E}\left[x|\mathcal{F}\right]$ the
expectation of a random variable $x$ with $\mathcal{F}$ denoting the
sigma algebra, which will be omitted when clear from the context.

%
%
\section{Problem Formulation}
We  consider a network of $m$ agents which solve the following
optimization problem cooperatively:
\begin{equation}\label{eq:optimization_formulation1}
\min\limits_{x\in\mathbb{R}^d} F(x)= \frac{1}{m}\sum_{i=1}^m
f_i(x),\quad f_i(x)\triangleq\mathbb{E}_{\xi_i\sim
\mathcal{D}_i}\left[\ell_i(x,\xi_i)\right]
\end{equation}
where $x$ is  common to all agents but $\ell_i:\mathbb{R}^d\times
\mathbb{R}\rightarrow\mathbb{R}$ is a local cost function
private to agent $i$. $\mathcal{D}_i$ is the local distribution of
random variable $\xi_i$ which usually denotes a data sample in
machine learning.

The above formulation also covers the empirical risk minimization
problem where $f_i(x)$ will be determined by
$
f_i(x)=\frac{1}{n_i}\sum_{j=1}^{n_i}\ell(x,\xi_{i,j})
$
with $n_i$ denoting  the number of data samples available to agent
$i$ and $\xi_{i,j}$ representing the randomness associated with the
$j$th sample for agent $i$.

Because of the randomness in $\ell_i(x,\xi_i)$, the gradient that each
agent $i$ can obtain is subject to noises. We denote the gradient
that agent $i$ has  for optimization at iteration $k$ as
{$g_i^k(x_i,\xi_i)$, with $x_i$ denoting the local copy
of the optimization variable on agent $i$. For the sake of
notational simplicity, we  will hereafter  abbreviate
$g_i^k(x_i,\xi_i)$  as $g_i^k$}. In this paper, we make the
following standard assumption about $f_i(\cdot)$ and $g_i^k$:
\begin{Assumption 1}\label{Assumption:gradient}
\begin{enumerate}
 \item {For any $1\leq i\leq m$, $x\in\mathbb{R}^d$, and $y\in\mathbb{R}^d$, we have
  $ \|\nabla f_i(x)-\nabla
f_i(y)\|\leq L
  \|x-y\|$  for some $L>0$}.
  \item All $g_i^k$ satisfy
$
  \mathbb{E}_{\xi_i\backsim\mathcal{D}_i}\left[g_i^k(x_i,\xi_i)\right]=\nabla f_i(x_i)
  $ and  $\mathbb{E}_{\xi_i\backsim\mathcal{D}_i}\left[\|g_i^k(x_i,\xi_i)-\nabla
  f_i(x_i)\|^2\right]\leq \sigma_g^2
  $ for any $x\in\mathbb{R}^d$.
\end{enumerate}
\end{Assumption 1}

We assume that the network of $m$ agents interact on an undirected
graph. The interaction can be described by a  matrix $W$. More
specifically, if agents $i$ and   $j$ can communicate and interact
with each other, then the $(i,j)$th entry of $W$, i.e., $w_{ij}$, is
positive. Otherwise, $w_{ij}$ is zero. Because every agent is able
to incorporate its own information in its optimization iteration
update, we have $w_{ii}>0$ for all $1\leq i\leq m$. The neighbor set
$\mathcal{N}_i$ of agent $i$ is defined as the set of agents
$\{j|w_{ij}>0\}$. So the neighbor set of agent $i$ always includes
itself. To ensure that the network can cooperatively solve
(\ref{eq:optimization_formulation1}), we make the following standard
assumption about the interaction: {\begin{Assumption
2}\label{assumption:coupling}
   The coupling matrix $W$ is doubly-stochastic,
  and the spectral radius of the matrix $ W-\frac{{\bf 1}{\bf 1}^T}{m}$, denoted as $\rho$, is less than
  one.
\end{Assumption 2}}


{In decentralized stochastic optimization, gradients
are directly computed from raw data and hence  embed sensitive
information of these data. For example,  in sensor network based
 localization, the positions of sensor agents should be kept private  in sensitive (hostile) environments \cite{zhang2019admm,huang2015differentially}.
 In  decentralized-optimization based localization
 approaches, the  gradient is a linear function of sensor locations and   disclosing the gradient  of an agent
amounts to disclosing its  position \cite{zhang2019admm}. In machine
learning applications, gradients are directly calculated from and
embed information of sensitive training data \cite{zhu2019deep}.  If
disclosed, these gradients can be used by an adversary to reversely
recover the raw data used for training. In fact, even when gradients
are subject to observation/measurement noise, the DLG attacker in
\cite{zhu2019deep} has been shown to be able to recover raw training
data from shared gradients, pixel-wise accurate for images and
token-wise matching for texts.}

%
%

 { We consider two  adversaries in decentralized stochastic optimization, honest-but-curious adversaries which are participating agents, and
  external eavesdroppers
\cite{Goldreich_2}:
\begin{itemize}
\item \emph{Honest-but-curious attacks}  are attacks in which a participating
agent or multiple participating agents (colluding or not)
 follows all protocol steps correctly but is curious and
collects all received intermediate data in an attempt to learn the
sensitive information about other participating agents.

\item \emph{Eavesdropping attacks}  are attacks in which an external eavesdropper eavesdrops upon all communication channels to
intercept exchanged messages so as to learn sensitive information
about sending agents.
\end{itemize}

An honest-but-curious adversary (e.g., agent $i$) has access to the
internal state  $x_i$, which is unavailable to external
eavesdroppers. However, an eavesdropper has access to all shared
information in the network, whereas an honest-but-curious agent can
only access shared information that is destined to it.}

{Inspired by recent work on information-theoretic
privacy \cite{sankar2013utility,diaz2019robustness}, we measure
 privacy using conditional differential entropy, which characterizes the minimum
expected squared error of any adversary's estimator.   More
specifically, for a private random variable $X$ and adversary's
observation $Y$, we quantify privacy using the conditional
differential entropy $h(X|Y)$. From information theory, for any
adversary's estimator $\hat{X}(Y)$, the following relationship holds
\cite{cover1999elements}:
\begin{equation}\label{eq:conditional_entropy_basic}
\mathbb{E}\left[\left(X-\hat{X}(Y)\right)^2\right]\geq
\frac{e^{2h(X|Y)}}{2\pi e}
\end{equation}
}

\section{An inherently privacy-preserving decentralized stochastic gradient  algorithm}
Before presenting our privacy-preserving decentralized stochastic
gradient descent algorithm, we first discuss why conventional
decentralized stochastic gradient descent algorithms leak private
information about participating agents.

The most commonly used decentralized stochastic gradient descent
algorithm {takes the following form
\cite{nedic2009distributed}}:
\[
x_i^{k+1}=\sum_{j\in\mathcal{N}_i}w_{ij}x_j^k-\lambda^k g_i^k
\]
where $x_i^k$ denotes the decision variable maintained by agent $i$
at iteration $k$, and { $\lambda^k$ denotes the stepsize
at iteration $k$}. Note that our defined  neighbor set
$\mathcal{N}_i$ always includes agent $i$.

In this conventional decentralized stochastic gradient descent
algorithm, every agent $i$ has to share $x_i^k$ with all its
neighbors. Because an adversary, e.g., an external eavesdropper, has
access to all shared information,   the adversary can calculate the
gradient of every agent $i$ based on publicly known $W$ and
$\lambda^k$. (Note that conventional algorithms need
  $w_{ij}$ to be publicly known to ensure that a doubly-stochastic $W$
  can be established in a decentralized way.) Therefore, an adversary can infer the gradient of participating
agents implementing   conventional decentralized stochastic gradient
descent algorithms.

Motivated by this observation, we propose to obscure the gradient
information using time-varying heterogeneous stepsizes and an
additional random mixing coefficient $b_{ij}^k$:
\begin{equation}\label{eq:proposed_algorithm}
x_i^{k+1}=\sum_{j\in\mathcal{N}_i}v_{ij}^k,\quad {\rm where}\quad
v_{ij}^k =w_{ij}x_j^k-b_{ij}^k\Lambda_j^k g_j^k
\end{equation}
where $\Lambda_j^k\triangleq{\rm
diag}\{\lambda_{j1}^k,\,\lambda_{j2}^k,\,\cdots,\lambda_{jd}^k\}\in\mathbb{R}^d$ is the stepsize matrix.
The diagonal elements of matrix $\Lambda_j^k$ are statistically
independent random variables determined by agent $j$ and they have the same
mathematical expectation $\bar\lambda_{j}^k$ and  standard deviation $\sigma_j^k$. The coefficient $b_{ij}^k\geq 0$ is
also determined by agent $j$ randomly under the constraint
$\sum_{i\in\mathcal{N}_j}b_{ij}^k=1$. Note that this constraint can
be satisfied locally because agent $j$   chooses all $b_{ij}^k$ for
its neighbors $i\in\mathcal{N}_j$ before sending $v^k_{ij}$ to agent
$i$. { It is worth noting that different from
gradient-tracking based decentralized optimization approaches (e.g.,
\cite{pu2020distributed,shi2015extra,qu2017harnessing}) that share
both gradient information and the optimization variable, our
approach only shares one mixed variable, which is significant when
the dimension of the optimization variable is high  and
communication bandwidth is limited.}

Note that in our proposed algorithm (\ref{eq:proposed_algorithm}),
at iteration $k$,   agent $j$ determines $\Lambda_j^k$ and
$b_{ij}^k$ randomly, and only sends $v_{ij}^k$ to its neighbors
$i\in\mathcal{N}_j,\,i\neq j$. Because both $\Lambda_j^k$ and
$b_{ij}^k$ are time-varying and private to agent $j$, they can
obscure the gradient $g_{j}^k$ of agent $j$. It is worth noting that
even when the dimension of gradient is high, in, e.g,   machine
learning applications, the statistically independent random
variables in $\Lambda_j^k$ will be able to obscure individual
components of $g_j^k$ independently. { It is also worth
noting that even when  an
 adversary has access to all   information shared in the network, since $v_{jj}^k$ is not  shared by agent $j$,
the adversary still cannot use the constraint
$\sum_{i\in\mathcal{N}_j}b_{ij}^k=1$ on $b^k_{ij}$ to infer gradient
information. In fact, in this case, the adversary can   obtain
$\sum_{i\in N_j,i\neq
j}v^k_{ij}=(1-w_{jj})x^k_j-(1-b_{jj}^k)\Lambda^k_jg^k_j$. Since
$b_{jj}^k$ is unknown to the adversary, it can help obfuscate the
gradient $g_j^k$  (so does $\Lambda^k_j$) against the adversary.}

Augmenting the decision variables of all agents as
\[
x^k=\left[\begin{array}{cccc}(x_1^k)^T&(x_2^k)^T&\cdots&(x_m^k)^T\end{array}\right]^T\in\mathbb{R}^{md\times
1}
\] and representing $B^k$ as a matrix with the $(i,\,j)$th element
equal to $b_{ij}^k$, we can write the network dynamics of the
proposed decentralized stochastic gradient descent algorithm as
follows:

\begin{equation}\label{eq:dynamics}
x^{k+1}=\hat{W} x^k-\hat{B}^k\hat{\Lambda}^k g^k
\end{equation}
where
\[
\hat{W}=W\otimes I_d \in\mathbb{R}^{md\times md}, \quad
\hat{B}^k=B^k\otimes I_d \in\mathbb{R}^{md\times md},
\]
\[\hat{\Lambda}^k={\rm
diag}\{\Lambda_1^k,\,\Lambda_2^k,\,\cdots,\Lambda_m^k\}
\in\mathbb{R}^{md\times md},
\]
\[
\begin{aligned}
&g^k=\left[(g_1^k)^T,\,(g_2^k)^T,\,\cdots,(g_m^k)^T\right]^T\in
\mathbb{R}^{md\times1}.
\end{aligned}
\]
Here $\otimes$ denotes Kronecker product and $I_d$ denotes the identity
matrix of dimension $d$. According to our assumption that all
diagonal elements of $\Lambda_{i}^k$ have the same mathematical
expectation $\bar\lambda_i^k$ and standard deviation $\sigma_i^k$, we have
$\mathbb{E}\left[\Lambda_i^k\right]=\bar{\lambda}_i^k I_d$ and $ {\rm var} \left[\Lambda_i^k\right]=(\sigma_i^k)^2 I_d$.
Furthermore, {since $b^k_{ij}$ satisfies
$\sum_{i\in\mathcal{N}_j}b_{ij}^k=1$,  $B^k$ is always a
column-stochastic matrix}.

In the following two sections, we will analyze the convergence  of
the proposed algorithm (\ref{eq:proposed_algorithm}) under convex
objective functions and general non-convex objective functions,
respectively.

\section{Convergence analysis in the convex objective function case}
In this section, we rigorously prove that { when the aggregate
objective function $F(\cdot)$ is convex and  the optimal
solution set of (\ref{eq:optimization_formulation1}) is nonempty},
the proposed algorithm will guarantee that all agents converge to a
same optimal solution. {Note that only the aggregate objective function $F(\cdot)$ is assumed to be convex and we do not require all individual objective functions $f_i(\cdot)$ to be convex. } 

{

To analyze the convergence of the proposed algorithm, we first introduce the following lemma:
\begin{Lemma 1}\label{th-dsystem}(Theorem 1 in \cite{wang2022tailoring})
Let  $\{\bv^k\}\subset \mathbb{R}^d$
and $\{\bu^k\}\subset \mathbb{R}^p$ be random nonnegative
vector sequences, and $\{a^k\}$ and $\{b^k\}$ be random nonnegative scalar sequences   such that
\[
\mathbb{E}\left[\bv^{k+1}|\mathcal{F}^k\right]\le (V^k+a^k{\bf 1}{\bf1}^T)\bv^k +b^k{\bf 1} -H^k\bu^k,\quad \forall k\geq 0
\]
holds almost surely, where $\{V^k\}$ and $\{H^k\}$ are random sequences of
nonnegative matrices and
$\mathbb{E}\left[\bv^{k+1}|\mathcal{F}^k \right]$ denotes the conditional expectation given
 $\bv^\ell,\bu^\ell,a^\ell,b^\ell,V^\ell,H^\ell$ for $\ell=0,1,\ldots,k$.
Assume that $\{a^k\}$ and $\{b^k\}$ satisfy
$\sum_{k=0}^\infty a^k<\infty$ and $\sum_{k=0}^\infty b^k<\infty$ almost surely, and
that there exists a (deterministic) vector $\pi>0$ such that almost surely
\[\pi^T V^k\le \pi^T,\qquad \pi^TH^k\ge 0\qquad\forall k\geq 0\]
Then, we have
1) $\{\pi^T\bv^k\}$ converges almost surely to some random variable $\pi^T\bv\geq 0$; 2) $\{\bv^k\}$ is bounded almost surely, and
3) $\sum_{ k=0 }^\infty \pi^TH^k\bu^k<\infty$ holds almost surely.
\end{Lemma 1}

}

{
Based on the above lemma, we first prove the following general convergence results for distributed algorithms for problem (\ref{eq:optimization_formulation1}):
\begin{Theorem 2}\label{th-main_decreasing}
Assume that problem (1) has a solution. Suppose that a distributed algorithm generates sequences
$\{x_i^k\}\subseteq\R^d$ such that
almost surely we have for any optimal solution $\theta^*$ of (\ref{eq:optimization_formulation1}),
\begin{equation}\label{eq:Theorem_decreasing}
\begin{aligned}
&\left[\begin{array}{c}
\mathbb{E}\left[\|\bar x^{k+1}-\theta^*\|^2|\mathcal{F}^k\right]\cr
\mathbb{E}\left[\sum_{i=1}^m\|x_i^{k+1}-\bar x^{k+1}\|^2|\mathcal{F}^k\right]\end{array}
\right]
\\
&\le \left( \left[\begin{array}{cc}
1 & \frac{1}{m}\cr
0& \rho\cr
\end{array}\right]
+a^k {\bf 1}{\bf 1}^T\right)\left[\begin{array}{c}\|\bar x^k-\theta^*\|^2\cr
\sum_{i=1}^m\|x_i^k-\bar x^k\|^2\end{array}\right]&&\cr
&\quad+b^k{\bf 1} - c^k \left[\begin{array}{c}
 F(\bar x^k)- F(\theta^*) \cr
 0\end{array}\right],\quad\forall k\geq 0
 \end{aligned}
\end{equation}
where  $\bar{x}^k=\frac{1}{m}\sum_{i=1}^m x_i^k$,
$\mathcal{F}^k=\{x_i^\ell, \, i\in[m],\, 0\le \ell\le k\}$,
the random
nonnegative scalar sequences $\{a^k\}$, $\{b^k\}$ satisfy $\sum_{k=0}^\infty a^k<\infty$ and $\sum_{k=0}^\infty b^k<\infty$ almost surely,  the deterministic  nonnegative sequence  $\{c^k\}$  satisfies
$
\sum_{k=0}^\infty c^k=\infty$, and $\rho$ satisfies $0<\rho<1$.
Then, we have
$\lim_{k\to\infty}\|x_i^k - \bar x^k\|=0$ almost surely for all agents $i$,
and there is an optimal solution $\tilde\theta^*$ of (\ref{eq:optimization_formulation1}) such that
$\lim_{k\to\infty}\|\bar x^k-\tilde\theta^*\|=0$ holds almost surely.
\end{Theorem 2}
\begin{proof}
Let $\theta^*$ be an arbitrary but fixed optimal solution of problem (1). Then, we have
 $F(\bar x^k)- F(\theta^*)\ge0$ for all $k$.
Hence,  by letting $\bv^k=\left[\|\bar x^k-\theta^*\|^2,\ \sum_{i=1}^m \|x_i^k-\bar x^k\|^2\right]^T$,
from relation~(\ref{eq:Theorem_decreasing}) it follows that almost surely for all $k\ge0$,
\be\label{eq-fin0}
\mathbb{E}\left[\bv^{k+1}|\mathcal{F}^k\right]
\le \left( \left[\begin{array}{cc}
1 & \frac{1}{m}\cr
0& \rho\cr\end{array}\right] +a^k {\bf 1}{\bf 1}^T \right)\bv^k+b^k{\bf 1}
\ee
Consider the vector $\pi=[1, \frac{1}{1-\rho}]^T$ and note
\[
\pi^T \left[\begin{array}{cc}
1 & \frac{1}{m}\cr
0& \rho\cr\end{array}\right]\leq \pi^T
\]

Thus, relation~\eqref{eq-fin0} satisfies all the conditions of Lemma~\ref{th-dsystem}.
By Lemma~\ref{th-dsystem}, it follows that
the limit $\lim_{k\to\infty}\pi^T\bv^k$ exists almost surely, and that the sequences $\{\|\bar x^k-\theta^*\|^2\}$
and $\{\sum_{i=1}^m \|x^k_i-\bar x^k\|^2\}$ are bounded almost surely.
From \eqref{eq-fin0} we have the following relation almost surely  for the second element of $\bv^k$:
\begin{equation*}
\begin{aligned}
&\mathbb{E}\left[\sum_{i=1}^m \|x_i^{k+1}-\bar x^{k+1}\|^2|\mathcal{F}^k\right]\cr
&\qquad\qquad\le  (\rho+ a^k)\sum_{i=1}^m \|x_i^k - \bar x^k\|^2 + \beta^k\quad\forall k\ge0
\end{aligned}
\end{equation*}
where
\[
\beta^k=a^k\left( \|\bar x^k - \theta^\ast\|^2+b^k\right)
\]
 Since
$\sum_{k=0}^\infty a^k<\infty$ holds almost surely by our assumption, and
the sequences $\{\|\bar x^k-\theta^*\|^2\}$
and $\{\sum_{i=1}^m \|x^k_i-\bar x^k\|^2\}$ are bounded almost surely, it follows that $\sum_{k=0}^\infty\beta^k<\infty$ holds almost surely.
Thus, the preceding relation satisfies the conditions of
Lemma~\ref{Lemma-polyak_2} in the appendix with
$v^k= \sum_{i=1}^m \|x_i^k - \bar x^k\|^2$, $\alpha^k=a^k$, $q^k=1-\rho$, and $p^k=\beta^k$
due to our assumptions that $\sum_{k=0}^\infty b^k<\infty$ holds almost surely and
$0<\rho<1$.
By Lemma \ref{Lemma-polyak_2} in the appendix it follows that we have the following relations almost surely:
\be\label{eq-sumable}
 \sum_{k=0}^\infty \sum_{i=1}^m \|x_i^k - \bar x^k\|^2<\infty,\:
 \lim_{k\to\infty} \sum_{i=1}^m \|x_i^k - \bar x^k\|^2=0
 \ee

 It remains to show that $\|\bar x^k-\theta^*\|^2\to0$ holds almost surely.
For this,
we consider relation~\eqref{eq:Theorem_decreasing}
and focus on the first element of $\bv^k$,  for which we obtain almost surely for all $k\ge0$:
\begin{equation}\label{eq:x_opt}
\begin{aligned}
&\mathbb{E}\left[\|\bar x^{k+1}-\theta^*\|^2|\mathcal{F}^k\right]
\le  (1+a^k)\|\bar x^k -\theta^*\|^2 \cr
&\quad+\left(\frac{1}{m}+a^k\right) \sum_{i=1}^m \|x_i^k - \bar x^k\|^2 + b^k -c^k  (F(\bar x^k)- F(\theta^*))
\end{aligned}
\end{equation}
The preceding relation satisfies Lemma~\ref{lem-opt} in the appendix
with $\phi=F$, $z^*=\theta^*$,
$z^k =\bar x^k$, $\a^k=a^k$, $\eta^k= c^k$, and
$\beta^k=
(\frac{1}{m}+a^k)
\sum_{i=1}^m \|x_i^k - \bar x^k\|^2+b^k$.
By our assumptions, the sequences $\{a^k\}$ and $\{b^k\}$ are summable almost surely, and $\sum_{k=0}^\infty c^k=\infty$. In view of~\eqref{eq-sumable}, it follows that
$\sum_{k=0}^\infty\beta^k<\infty$ holds almost surely.
Hence, all the conditions of Lemma~\ref{lem-opt} in the appendix
are satisfied and, consequently, $\{\bar x^k\}$ converges to some optimal solution almost surely.
\end{proof}

}

Having obtained Theorem \ref{th-main_decreasing}, we are in
position to prove that all agents will converge  to a same optimal
solution of (\ref{eq:optimization_formulation1}) when the aggregate objective function
$F(\cdot)$ is convex.
\begin{Theorem 1}\label{Theorem:Theorem_1}
Under Assumption \ref{Assumption:gradient} and Assumption
\ref{assumption:coupling}, if the aggregate objective function $F(\cdot)$ is convex and the
optimal solution set of problem (\ref{eq:optimization_formulation1})
is non-empty, then all $x_i^k$ $(1\leq p \leq m)$ converge to a  solution
 of problem (\ref{eq:optimization_formulation1}) if the mathematical expectation
and standard deviation of stepsize $\Lambda_i^k$ (represented as $\bar\lambda_i^kI_d$ and  $\sigma_i^kI_d$, respectively) satisfy for all $i$:
  \begin{enumerate}
  \item  non-summable but square summable condition:
\begin{equation}\label{eq:non_summable}
  \sum_{k=0}^{\infty}\bar\lambda_i^k=+\infty, \quad
\sum_{k=0}^{\infty}(\bar\lambda_i^k)^2<\infty, \quad
\sum_{k=0}^{\infty}(\sigma_i^k)^2<\infty
\end{equation}
  \item  heterogeneity condition:
\begin{equation}\label{eq:heterogeneity_condition}
 {\sum_{k=0}^{\infty}\: \sum_{i,j\in\{1,2,\cdots,m\},\,i\neq
j}|\bar\lambda_i^{k}-\bar\lambda_j^k|}<\infty
\end{equation}
  \end{enumerate}

\end{Theorem 1}

{
\begin{proof}
The basic idea is to apply Theorem \ref{th-main_decreasing}. We establish needed relationships  for  $\mathbb{E}\left[\|\bar x^{k+1}-\theta^*\|^2|\mathcal{F}^k\right]$ and $\mathbb{E}\left[\sum_{i=1}^m\|x_i^{k+1}-\bar x^{k+1}\|^2|\mathcal{F}^k\right]$ in Step I and Step II below, respectively.

Part I: We first analyze  $\|\bar x^{k+1}-\theta^*\|^2$.

Using the definition $\bar{x}^k=\frac{1}{m}\sum_{i=1}^m x_i^k$
and (\ref{eq:dynamics}), we have
\begin{equation}\label{eq:x_bar_dynamics}
\begin{aligned}
\bar{x}^{k+1}&=\frac{{\bf 1}^T\otimes I_d}{m} x^{k+1}\\
&= \frac{{\bf 1}^T\otimes I_d}{m}\hat{W} x^k-\frac{{\bf 1}^T\otimes I_d}{m}\hat{B}^k\hat{\Lambda}^k g^k\\
&=\bar{x}^{k}+\frac{{\bf 1}^T\otimes I_d}{m}\hat{B}^k\hat{\Lambda}^k g^k\\
&=\bar{x}^{k}- \frac{1}{m} {\sum_{i=1}^m\Lambda_i^k g_i^k}
\end{aligned}
\end{equation}
where in the third equality we used the doubly-stochastic property of $W$, and in the last equality we used the column stochastic property of $B^k$. Therefore, for any optimal solution $\theta^*$, we have
\begin{equation}\label{eq:x_bar_k+1}
\bar x^{k+1}-\theta^*=\bar x^k -\theta^* -  \frac{1}{m} {\sum_{i=1}^m\Lambda_i^k g_i^k}
\end{equation}
The preceding relation implies
\begin{equation}\label{eq-rel1}
\begin{aligned}
 &\left\|\bar x^{k+1}-\theta^*\right\|^2\\
 &=\left\|\bar
x^k -\theta^* \right\|^2 -  \frac{2}{m} \sum_{i=1}^m
\left\langle \Lambda_i^kg_i^k, \bar x^k -\theta^*
\right\rangle   +  \frac{1}{m^2} \left\|\sum_{i=1}^m\Lambda_i^k
 g_i^k\right\|^2\\
 &= \left\|\bar
x^k -\theta^* \right\|^2 -  \frac{2}{m} \sum_{i=1}^m
\left\langle \Lambda_i^kg_i^k, \bar x^k -\theta^*
\right\rangle   + \frac{1}{m^2}\|\Lambda^k g^k\|^2\\
  &\leq \left\|\bar
x^k -\theta^* \right\|^2 -  \frac{2}{m} \sum_{i=1}^m
\left\langle \Lambda_i^kg_i^k, \bar x^k -\theta^*
\right\rangle  + \frac{1}{m^2}\|\Lambda^k\|^2\|g^k\|^2
\end{aligned}
\end{equation}
where we defined  $ \Lambda^k=[\Lambda_1^k,\cdots,\Lambda_m^k]^T\in\mathbb{R}^{d\times md}$.

Defining
\begin{equation}\label{eq:nabla_f1}
\nabla f(x^k)=\left[(\nabla f_1(x_1^k))^T,\cdots, (\nabla f_m(x_m^k))^T\right]^T
\end{equation}
we have

\begin{equation}\label{eq:gradient}
\begin{aligned}
\|g^k\|^2 &= \|g^k-\nabla f(x^k)+\nabla f(x^k)\|^2 \\
& \leq 2 \|g^k-\nabla f(x^k)\|^2+2\|\nabla f(x^k)\|^2
\end{aligned}
\end{equation}

Denoting $x^\ast={\bf 1}\otimes\theta^\ast$ and then adding and subtracting $\nabla f(x^\ast)$ to $\nabla f(x^k)$ yield
\begin{equation}\label{eq:add_subtract}
\begin{aligned}
  \|\nabla f(x^k)\|^2&=\|\nabla f(x^k)-\nabla f(x^\ast)+\nabla f(x^\ast)\|^2\\
  &\leq 2\|\nabla f(x^k)-\nabla f(x^\ast)\|^2+2\|\nabla f(x^\ast)\|^2\\
  &\leq 2L^2\|x^k- x^\ast\|^2+2\|\nabla f(x^\ast)\|^2\\
  &\leq 2L^2\sum_{i=1}^{m}\|x_i^k- \theta^\ast\|^2+2\|\nabla f(x^\ast)\|^2\\
   &\leq 2L^2\sum_{i=1}^{m}\|x_i^k- \bar{x}^k+\bar{x}^k-\theta^\ast\|^2+2\|\nabla f(x^\ast)\|^2\\
  &\leq 4L^2\sum_{i=1}^{m}\|x_i^k- \bar{x}^k\|^2+4mL^2 \|  \bar{x}^k- \theta^\ast\|^2\\
  &\qquad+2\|\nabla f(x^\ast)\|^2
  \end{aligned}
\end{equation}

Combining (\ref{eq-rel1}), (\ref{eq:gradient}), and (\ref{eq:add_subtract})   yields
\begin{equation}\label{eq-rel2}
\begin{aligned}
 &\left\|\bar x^{k+1}-\theta^*\right\|^2\\
 &\leq \left\|\bar
x^k -\theta^* \right\|^2 -  \frac{2}{m} \sum_{i=1}^m
\left\langle \Lambda_i^kg_i^k, \bar x^k -\theta^*
\right\rangle +\frac{\|\Lambda^k\|^2}{m^2}\times\\
 &\qquad \bigg(8L^2\sum_{i=1}^{m}\|x_i^k- \bar{x}^k\|^2+8mL^2 \|  \bar{x}^k- \theta^\ast\|^2\\
 & \qquad+4\|\nabla f(x^\ast)\|^2+2 \|g^k-\nabla f(x^k)\|^2\bigg)
\end{aligned}
\end{equation}

Taking the conditional expectation, given $\mathcal{F}^k=\{x^0,\ldots,x^k\}$, and
using the  relationship $\| \Lambda^k\|^2\leq \| \Lambda^k\|_F^2=\sum_{j=1}^{m}\sum_{p=1}^{d}(\lambda_{jp}^k)^2$, we obtain
\begin{equation}\label{eq-rel3}
\begin{aligned}
 &\mathbb{E}\left[\left\|\bar x^{k+1}-\theta^*\right\|^2|\mathcal{F}^k\right]  \\
&\leq\left\|\bar
x^k -\theta^* \right\|^2-  \frac{2}{m} \sum_{i=1}^m
\left\langle \bar\lambda_i^k\nabla f_i(x_i^k), \bar x^k -\theta^*
\right\rangle\\
 &\qquad+\frac{\sum_{i=1}^{m} d\left((\bar\lambda_i^k)^2+(\sigma_i^k)^2\right)}{m^2} \times\bigg(8L^2\sum_{i=1}^{m}\|x_i^k- \bar{x}^k\|^2\\
 & \qquad+8mL^2 \|  \bar{x}^k- \theta^\ast\|^2+4\|\nabla f(x^\ast)\|^2+2m\sigma_g^2\bigg)
\end{aligned}
\end{equation}
Note that $\sigma_g^2$ is the variance of stochastic gradients  from Assumption \ref{Assumption:gradient}, and we used the fact that the mathematical expectation and standard deviation of  $ \Lambda_i^k$ is $\bar{\lambda}_i^k I_d$ and $\sigma_i^k I_d$, respectively.

 We next estimate the inner-product term, for which we have
\begin{equation}\label{eq:inner_product_0}
\begin{aligned}
 \left\langle \bar\lambda_i^k \nabla f_i(x_i^k), \bar x^k -\theta^* \right\rangle= &\left\langle \bar\lambda_i^k(\nabla f_i(x_i^k)-\nabla f_i(\bar{x}^k)),  \bar x^k -\theta^*
 \right\rangle\\
 &+\left\langle\bar\lambda_i^k\nabla f_i(\bar{x}^k), \bar x^k -\theta^*
 \right\rangle
 \end{aligned}
\end{equation}

Using the
Lipschitz  property of $\nabla f_i$ in Assumption \ref{Assumption:gradient}, we have
\begin{equation}\label{eq:inner_product_1}
\begin{aligned}
&\left\langle \bar\lambda_i^k(\nabla f_i(x_i^k)-\nabla f_i(\bar{x}^k)),  \bar x^k
-\theta^*
 \right\rangle\\
 &\qquad\geq -L\bar\lambda_i^k\|x_i^k-\bar{x}^k\|\|\bar x^k -\theta^*\|\\
&\qquad\geq- \frac{1}{2}\|x_i^k-\bar{x}^k\|^2
- \frac{1}{2}L^2(\bar\lambda_i^k)^2\|\bar x^k -\theta^*\|^2
\end{aligned}
\end{equation}
 Defining  $\bar\lambda^k\triangleq \frac{\sum_{i=1}^{m}\bar\lambda_i^k}{m}$, we have
 \begin{equation}\label{eq:inner_product_2}
\begin{aligned}
&\left\langle\bar\lambda_i^k\nabla f_i(\bar{x}^k), \bar x^k
-\theta^*
 \right\rangle=\\
  &\qquad \left\langle(\bar\lambda_i^k-\bar\lambda^k)\nabla
f_i(\bar{x}^k), \bar x^k -\theta^*
 \right\rangle +\left\langle\bar\lambda^k\nabla f_i(\bar{x}^k), \bar x^k -\theta^*
 \right\rangle
 \end{aligned}
\end{equation}

 Further defining a vector  $\bar{\pmb\lambda}^k = [\bar\lambda_1^k, \cdots,\bar\lambda_m^k]^T$ and
combining (\ref{eq:inner_product_0})-(\ref{eq:inner_product_2})
yield
\begin{equation}\label{eq:inner_product_3}
\begin{aligned}
&\frac{\sum_{i=1}^m \left\langle \bar\lambda_i^k\nabla f_i(x_i^k), \bar x^k -\theta^*
\right\rangle}{m}\\
&\geq \frac{\sum_{i=1}^m (-\|x_i^k-\bar{x}^k\|^2
-L^2(\bar\lambda_i^k)^2\|\bar x^k -\theta^*\|^2)}{2m}+\\
&\frac{\sum_{i=1}^m\left(\left\langle(\bar\lambda_i^k-\bar\lambda^k)\nabla
f_i(\bar{x}^k), \bar x^k -\theta^*
 \right\rangle +\left\langle\bar\lambda^k\nabla f_i(\bar{x}^k), \bar x^k -\theta^*
 \right\rangle\right)}{m}\\
 &= -\frac{\sum_{i=1}^m  \|x_i^k-\bar{x}^k\|^2
}{2m}-\frac{L^2\|\bar{\pmb\lambda}^k\|^2\|\bar x^k -\theta^*\|^2 }{2m}+\\
&\frac{\sum_{i=1}^m \left\langle(\bar\lambda_i^k-\bar\lambda^k)\nabla
f_i(\bar{x}^k), \bar x^k -\theta^*
 \right\rangle }{m}+\bar\lambda^k\left\langle\nabla F(\bar x^k), \bar x^k
 -\theta^*\right\rangle\\
 &\geq -\frac{\sum_{i=1}^m  \|x_i^k-\bar{x}^k\|^2
}{2m}-\frac{L^2\|\bar{\pmb\lambda}^k\|^2\|\bar x^k -\theta^*\|^2 }{2m}+\\
&\frac{\sum_{i=1}^m \left\langle(\bar\lambda_i^k-\bar\lambda^k)\nabla
f_i(\bar{x}^k), \bar x^k -\theta^*
 \right\rangle }{m}+\bar\lambda^k(F(\bar x^k)-F(\theta^*)
\end{aligned}
\end{equation}
where we used the convexity of $F(\cdot)$ in the last inequality.

Noting  $\bar{\pmb\lambda}^k=[\bar\lambda_1^k,\cdots,\bar\lambda_m^k]^T$ and $\bar\lambda^k=\frac{\sum_{i=1}^{m}\bar\lambda_i^k}{m}$, we always have
\begin{equation}\label{eq:inequality_in_Theo1}
\begin{aligned}
&\frac{\sum_{i=1}^m \left\langle(\bar\lambda_i^k-\bar\lambda^k)\nabla
f_i(\bar{x}^k), \bar x^k -\theta^*
 \right\rangle }{m}\\
 &=\frac{ \left\langle\sum_{i=1}^m(\bar\lambda_i^k-\bar\lambda^k)\nabla
f_i(\bar{x}^k), \bar x^k -\theta^*
 \right\rangle }{m}\\
 &\geq - \frac{ \|\sum_{i=1}^m(\bar\lambda_i^k-\bar\lambda^k)\nabla
f_i(\bar{x}^k)\|\: \|\bar x^k -\theta^*
 \|}{m}\\
 &=- \frac{\| \left((\bar{\pmb\lambda}^k-\bar\lambda^k{\bf 1}_m)\otimes {\bf 1}_d\right)^T \nabla
f( {\bf 1}\otimes\bar{x}^k)\|\: \|\bar x^k -\theta^*
 \|}{m}\\
 &\geq - \frac{\sqrt{d}}{m}\|  \bar{\pmb\lambda}^k-\bar\lambda^k{\bf 1}_m\| \:\|\nabla
f({\bf 1}\otimes\bar{x}^k)\|\: \|\bar x^k -\theta^*
 \|\\
 &\geq - \frac{\sqrt{d}}{2m}\|  \bar{\pmb\lambda}^k-\bar\lambda^k{\bf 1}_m\| \left(\|\nabla
f({\bf 1}\otimes\bar{x}^k)\|^2 +\|\bar x^k -\theta^*
 \|^2\right)
\end{aligned}
\end{equation}
where $\otimes$ is Kronecker product and $\nabla f$ is defined in (\ref{eq:nabla_f1}).

Furthermore, $\|\nabla f({\bf
1}\otimes\bar x^k)\|$ can be bounded by
\begin{equation}\label{eq:nabla_f}
\begin{aligned}
\|\nabla f({\bf 1}\otimes\bar x^k)\|&= \|\nabla f({\bf 1}\otimes\bar
x^k)-\nabla f(x^*)+\nabla f(x^*)\|\\
&\leq \|\nabla f({\bf 1}\otimes\bar
x^k)-\nabla f(x^*)\|+\|\nabla f(x^*)\|
\\
&\le L\|{\bf 1}\otimes\bar x^k- x^*\|+\|\nabla f(x^*)\|\\
&=L\sqrt{m}\, \|\bar x^k -
\theta^*\|+\|\nabla f(x^*)\|
\end{aligned}
\end{equation}

 Combining (\ref{eq:inner_product_3}),
(\ref{eq:inequality_in_Theo1}), and (\ref{eq:nabla_f})
 yields
\begin{equation}\label{eq:inner_product_4}
\begin{aligned}
&\frac{\sum_{i=1}^m \left\langle \bar\lambda_i^kg_i^k, \bar
x^k -\theta^*
\right\rangle}{m}\\
  &\geq -\frac{\sum_{i=1}^m  \|x_i^k-\bar{x}^k\|^2
}{2m}-\frac{L^2\|\bar{\pmb\lambda}^k\|^2\|\bar x^k -\theta^*\|^2 }{2m}\\
 &\quad - \frac{\sqrt{d}(2L^2m+1)}{2m}\|  \bar{\pmb\lambda}^k-\bar\lambda^k{\bf 1}_m\|
 \|\bar x^k -\theta^*
 \|^2     \\
 &\quad -  \frac{\sqrt{d}}{m}\|  \bar{\pmb\lambda}^k-\bar\lambda^k{\bf 1}_m\|  \|\nabla f(x^*)\|^2+\bar\lambda^k(F(\bar x^k)-F(\theta^*)
\end{aligned}
\end{equation}

 Substituting (\ref{eq:inner_product_4}) into (\ref{eq-rel3}) leads to
 \begin{equation}\label{eq:bar_x-theta}
\begin{aligned}
 &\mathbb{E}\left[\left\|\bar x^{k+1}-\theta^*\right\|^2|\mathcal{F}^k\right]\\
 &\leq \left\|\bar
x^k -\theta^* \right\|^2
+\frac{\sum_{i=1}^m  \|x_i^k-\bar{x}^k\|^2
}{m}+\frac{L^2\|\bar{\pmb\lambda}^k\|^2\|\bar x^k -\theta^*\|^2 }{m}\\
 &\quad  +\frac{\sqrt{d}(2L^2m+1)}{m}\|  \bar{\pmb\lambda}^k-\bar\lambda^k{\bf 1}_m\|
 \|\bar x^k -\theta^*
 \|^2   \\
 &\quad + \frac{2\sqrt{d}}{m} \|  \bar{\pmb\lambda}^k-\bar\lambda^k{\bf 1}_m\|  \|\nabla f(x^*)\|^2-2\bar\lambda^k(F(\bar x^k)-F(\theta^*)
 \\
 &\quad+\frac{\sum_{i=1}^{m} d\left((\bar\lambda_i^k)^2+(\sigma_i^k)^2\right)}{m^2} \times\bigg(8L^2\sum_{i=1}^{m}\|x_i^k- \bar{x}^k\|^2\\
 & \qquad+8mL^2 \|  \bar{x}^k- \theta^\ast\|^2+4\|\nabla f(x^\ast)\|^2+2m\sigma_g^2\bigg)
\end{aligned}
\end{equation}

 Part II: Next we  analyze $\sum_{i=1}^m\|x_i^{k+1}-\bar x^{k+1}\|^2$.

 Using (\ref{eq:dynamics}) and (\ref{eq:x_bar_dynamics}), and defining $\hat\Pi=(I -\frac{\bf 1 {\bf 1 }^T}{m}) \otimes I_d $, we have
\end{proof}
\[
\begin{aligned}
x^{k+1}-{\bf 1}\otimes \bar x^{k+1} &=\hat{W}x^k -{\bf 1}\otimes \bar x^{k} -\hat\Pi\hat{B}^k\hat\Lambda^k g^k \\
&= \hat{W}(x^k -{\bf 1}\otimes \bar x^{k}) -\hat\Pi\hat{B}^k\hat\Lambda^k g^k
\end{aligned}
\]
 where we used the doubly-stochastic property of $W$ which leads to $\hat{W}({\bf 1}\otimes \bar x^{k})={\bf 1}\otimes \bar x^{k}$.

Taking  norm on both sides and using   $\rho=\|W- \frac{1}{m}{\bf
1}{\bf 1}^T\|$ from Assumption \ref{assumption:coupling}, we obtain

\[
\begin{aligned}
\|x^{k+1}-{\bf 1}\otimes &\bar x^{k+1}\| \\
&\leq \rho  \|x^{k}-{\bf 1}\otimes \bar x^{k}\|+\|\hat\Pi\| \|\hat{B}^k\| \|\hat\Lambda^k\| \|g^k\|\\
& \leq \rho  \|x^{k}-{\bf 1}\otimes \bar x^{k}\|+  md\|\hat\Lambda^k\| \|g^k\|
\end{aligned}
\]
where in the  second inequality we used $\|\Pi\|=1$ and $ \|\hat B^k \|\le  \|\hat B^k \|_F  \leq md$ with $\|\cdot\|_F$ denoting the Frobenius matrix norm.

By taking squares on both sides  and using the inequality $2ab\le
\epsilon a^2 + \epsilon^{-1}b^2$ valid for any  $a\in\mathbb{R}$, $b\in\mathbb{R}$, and
$\epsilon>0$, we obtain (noting $\|x^{k}-{\bf 1}\otimes \bar x^{k}\|^2=\sum_{i=1}^{m}\|x_i^k-\bar{x}^k\|^2$):

\begin{equation}\label{eq:x^{k+1}-barx_norm}
\begin{aligned}
\sum_{i=1}^m\|x^{k+1}_i - \bar x^{k+1} \|^2\le&
\rho^2(1+\epsilon)\sum_{i=1}^m\|x^k_i - \bar x^k \|^2 \\
& +m^2d^2(1+\epsilon^{-1})\|\hat\Lambda^k\|^2\|g^k\|^2
\end{aligned}
\end{equation}
By letting $1+\epsilon=\frac{1}{\rho}$  and hence
$1+\epsilon^{-1}=(1-\rho)^{-1}$,  we arrive at
\begin{equation}\label{eq:x^{k+1}-barx_norm1}
\begin{aligned}
\sum_{i=1}^m\|x^{k+1}_i - \bar x^{k+1} \|^2\le&
\rho \sum_{i=1}^m\|x^k_i - \bar x^k \|^2  +\frac{m^2d^2}{1-\rho}\|\hat\Lambda^k\|^2\|g^k\|^2
\end{aligned}
\end{equation}

Noting $\|\hat\Lambda^k\|=\max\limits_{1\leq j\leq m, 1\leq p\leq d}\lambda_{jp}^k$, and using the relationship (\ref{eq:gradient}) and (\ref{eq:add_subtract}) for $\|g^k\|$, we can obtain
\begin{equation}\label{eq:x^{k+1}-barx_norm2}
\begin{aligned}
\sum_{i=1}^m&\|x^{k+1}_i - \bar x^{k+1} \|^2\\
&\le
\rho \sum_{i=1}^m\|x^k_i - \bar x^k \|^2+\frac{m^2d^2\max\limits_{j,p}(\lambda_{jp}^k)^2}{1-\rho} \times\\
&\quad\bigg(8L^2\sum_{i=1}^{m}\|x_i^k- \bar{x}^k\|^2+8mL^2 \|  \bar{x}^k- \theta^\ast\|^2\\
 & \qquad\qquad+4\|\nabla f(x^\ast)\|^2+2 \|g^k-\nabla f(x^k)\|^2\bigg)
\end{aligned}
\end{equation}
Taking the conditional expectation, given $\mathcal{F}^k=\{x^0,\ldots,x^k\}$ yields
\begin{equation}\label{eq:x^{k+1}-barx_norm4}
\begin{aligned}
\mathbb{E}&\left[\sum_{i=1}^m\|x^{k+1}_i - \bar x^{k+1} \|^2\right]\\
&\le
\rho \sum_{i=1}^m\|x^k_i - \bar x^k \|^2  +\frac{m^2d^2\max\limits_{i}\left((\bar\lambda_i^k)^2+(\sigma_i^k)^2\right)}{1-\rho} \times\\
&\\
&\quad\bigg(8L^2\sum_{i=1}^{m}\|x_i^k- \bar{x}^k\|^2+8mL^2 \|  \bar{x}^k- \theta^\ast\|^2\\
 & \qquad\qquad+4\|\nabla f(x^\ast)\|^2+2m\sigma_g^2\bigg)
\end{aligned}
\end{equation}

Combining (\ref{eq:bar_x-theta}) and (\ref{eq:x^{k+1}-barx_norm4}),
we have
\begin{equation}\label{algorithm_proof_final}
\mathbb{E}\left[\bv^{k+1}|\mathcal{F}^k\right] \le \left( \left[\begin{array}{cc} 1 & \frac{1}{m}\cr 0&
\rho\cr
\end{array}\right]
+A^k \right)\bv^k  +{\bf b}^k- 2\bar\lambda^k{\bf u}^k
\end{equation}
where
\[
\bv^k=\left[\begin{array}{c}\|\bar x^k-\theta^*\|^2\cr
\sum_{i=1}^m\|x_i^k-\bar x^k\|^2\end{array}\right],
\]

\[
{\bf u}^k= \left[\begin{array}{c}
 F(\bar x^k)-F(\theta^*)\cr 0\end{array}\right]
,\quad
  A^k=
\left[\begin{array}{cc}A_{11}&A_{12}\\A_{21}&A_{22}\end{array}\right]
\] with
\[
\begin{aligned}
A_{11}=&  \frac{L^2\|\bar{\pmb\lambda}^k\|^2}{m}+ \frac{\sqrt{d}(2L^2m+1)}{m}\|  \bar{\pmb\lambda}^k-\bar\lambda^k{\bf 1}_m\|  \\
       & +\frac{8mL^2\sum_{i=1}^{m} d\left((\bar\lambda_i^k)^2+(\sigma_i^k)^2\right)}{m^2},\\
A_{12}&=\frac{8L^2\sum_{i=1}^{m} d\left((\bar\lambda_i^k)^2+(\sigma_i^k)^2\right)}{m^2}\\
A_{21}&=\frac{8m^3L^2d^2\max\limits_{i}\left((\bar\lambda_i^k)^2+(\sigma_i^k)^2\right)}{1-\rho}\\
A_{22}&=\frac{8m^2L^2d^2\max\limits_{i}\left((\bar\lambda_i^k)^2+(\sigma_i^k)^2\right)}{1-\rho}
\end{aligned}
\]
 and ${\bf b}^k=\left[\begin{array}{c} b_1^k\\b_2^k\end{array}\right]$ with
 \[
 \begin{aligned}
 b_1^k=&\frac{2\sqrt{d}}{m} \|\bar{\pmb\lambda}^k-\bar\lambda^k{\bf 1}_m\| \|\nabla f(x^*)\|^2 \\
 &+ \frac{\sum_{i=1}^{m} d\left((\bar\lambda_i^k)^2+(\sigma_i^k)^2\right)}{m^2}   ( 4\|\nabla f(x^\ast)\|^2+2m\sigma_g^2 ),\\
 b_2^k=& \frac{m^2d^2\max\limits_{i}\left((\bar\lambda_i^k)^2+(\sigma_i^k)^2\right)( 4\|\nabla f(x^\ast)\|^2+2m\sigma_g^2 )}{1-\rho}
 \end{aligned}
 \]

Because   $A^k\leq a^k {\bf 1}{\bf 1}^T $ and ${\bf b}^k\leq b^k{\bf 1}$ hold when
$a^k$ and $b^k$ are set to
$a^k=\max\left\{A_{11},A_{12},A_{21},A_{22}\right\}$ and   $b^k=\max\left\{b_{1}^k,b_{2}^k\right\}$, respectively, we can see
that (\ref{eq:Theorem_decreasing}) in Theorem \ref{th-main_decreasing} is satisfied. Further
note that under the conditions in the statement, all conditions for
$\{a^k\}$, $\{b^k\}$, and $\{c^k\}$ in Theorem \ref{th-main_decreasing} are
also satisfied. Therefore, we have the
claimed results.
}

\begin{Remark 1} In implementation, to satisfy the
heterogeneity condition in (\ref{eq:heterogeneity_condition}), all
agents can choose the same expected value for their stepsizes. For
example, all agents can set their expected stepsizes to
$\frac{1}{k}$, i.e.,
$\bar{\lambda}_1^k=\bar{\lambda}_2^k=\cdots=\bar{\lambda}_m^k=\frac{1}{k}$.
Since it is the exact random outcome  $\lambda_i^k$ that is used to
obfuscate the gradient, and the exact value $\lambda_i^k$ is known
only
 to agent
  $i$, the privacy of every agent $i$ can still be protected even when $\bar\lambda_i^k$ is publicly known. Moreover, individual agents can also  avoid revealing the expected values
  of their stepsizes. For example, every agent can  deviate  its expected stepsize  from the
baseline $\frac{1}{k}$ in a finite number of iterations. The indices
of these iterations are private to individual agents. As long as the
deviation in each of these iterations is  finite,  the heterogeneity
condition in (\ref{eq:heterogeneity_condition}) will still be
satisfied.
\end{Remark 1}
{ 
\begin{Remark 1}
Because the evolution of $x_i^k$ to the optimal solution satisfies the conditions in Theorem \ref{th-main_decreasing}, we can leverage Theorem \ref{th-main_decreasing} to examine the convergence speed.  The first relationship in (\ref{eq-sumable}) (i.e., $\sum_{k=0}^\infty \sum_{i=1}^m \|x_i^k - \bar x^k\|^2<\infty$)  implies that   $\sum_{i=1}^m \|x_i^k - \bar x^k\|^2$  decreases to zero with a rate no slower than $\mathcal{O}(\frac{1}{k})$, and hence we have $x_i^k$ converging to $\bar{x}^k$  no slower than $\mathcal{O}(\frac{1}{k^{0.5}})$. Moreover,  given that $a^k$ and $b^k$ in (\ref{eq:x_opt}) are summable (and hence decrease  to zero no slower than $\mathcal{O}(\frac{1}{k})$) and $c^k$ in (\ref{eq:x_opt})  corresponds to the average expected stepsize $\bar{\lambda}^k=\frac{\sum_{i=1}^{m}\bar\lambda_i^k}{m}$ in our algorithm (which is square summable and hence decreases to zero no slower than $\mathcal{O}(\frac{1}{k^{0.5}}))$, we have that $\bar{x}^k$ converges to an optimal solution with a speed no worse than $\mathcal{O}(\frac{1}{k^{0.5}})$ \cite{polyak87}. Therefore, the convergence of every $x_i^k$ to an optimal solution, which is equivalent to the combination of the convergence of $x_i^k$ to $\bar{x}^k$ and the convergence of $\bar{x}^k$ to an optimal solution,  should be no slower than $\mathcal{O}(\frac{1}{k^{0.5}})$.
\end{Remark 1}}
\section{Convergence analysis in the non-convex objective function case}
In this section, we show that the proposed algorithm will ensure
convergence of all agents   to a same stationary point  when the
objective functions are non-convex, as is the case in most machine
learning applications.

\begin{Theorem 1}\label{theorem_non_convex}
Under Assumption \ref{Assumption:gradient} and Assumption
\ref{assumption:coupling}, if $f_i(\cdot)$ $(1\leq i\leq m)$ can be non-convex but have bounded gradients, then when the non-summable but square summable
condition in (\ref{eq:non_summable}) and the heterogeneity condition
in (\ref{eq:heterogeneity_condition}) for stepsizes are satisfied,
we have that the proposed algorithm will guarantee   the following
results for all $1\leq p\leq m$:
\begin{equation}\label{eq:consensus}
\lim_{k\rightarrow\infty}\|x_p^k-\bar{x}^k\|=0,\quad  {\rm almost}\:{\rm surely}
\end{equation}
\begin{equation}\label{eq:gradient_summable1}
\lim_{t\rightarrow\infty}\frac{\sum_{k=0}^t
\bar\lambda_p^k\mathbb{E}\left[\left\|\nabla
F(\bar{x}^k)\right\|^2\right]}{\sum_{k=0}^t \bar{\lambda}_p^k}=0
\end{equation}
\begin{equation}\label{eq:graident_summable2}
\lim_{t\rightarrow\infty}\frac{\sum_{k=0}^t \bar\lambda_p^k
\mathbb{E}\left[\left\|\frac{\sum_{i=1}^m\nabla f_i(x_i^k)
}{m}\right\|^2\right]}{\sum_{k=0}^t\bar\lambda_p^k}=0
\end{equation}
\end{Theorem 1}

\begin{proof}
{ 
We first prove (\ref{eq:consensus}).

Since the derivation in the proof of Part II of Theorem \ref{Theorem:Theorem_1} is independent of the convexity of $F(\cdot)$, we still have
the relationship in (\ref{eq:x^{k+1}-barx_norm1}) for $\sum_{i=1}^m\|x^{k+1}_i - \bar x^{k+1} \|^2$, which further leads to
\begin{equation}
\begin{aligned}
&\sum_{i=1}^m\|x^{k+1}_i - \bar x^{k+1} \|^2\le
\rho \sum_{i=1}^m\|x^k_i - \bar x^k \|^2 \\
&\qquad\qquad +\frac{m^2d^2\max\limits_{j,p}(\lambda_{jp}^k)^2}{1-\rho}G^2
 \end{aligned}
\end{equation}
where we used $\|\hat\Lambda^k\|=\max\limits_{1\leq j\leq m, 1\leq p\leq d}\lambda_{jp}^k$  and that  $\|g^k\| $ is bounded (with the bound  represented by  $G>0$ here).

Taking the conditional expectation, given $\mathcal{F}^k=\{x^0,\ldots,x^k\}$ yields
\begin{equation}\label{eq:x^{k+1}-barx_norm3}
\begin{aligned}
\mathbb{E}\left[\sum_{i=1}^m\|x^{k+1}_i - \bar x^{k+1} \|^2\right]&\le
\rho \sum_{i=1}^m\|x^k_i - \bar x^k \|^2  \\
&+\frac{m^2d^2\max\limits_{i}\left((\bar\lambda_i^k)^2+(\sigma_i^k)^2\right)}{1-\rho}G^2
\end{aligned}
\end{equation}
Since $\rho<1$ holds and both $(\bar\lambda_i^k)^2$ and $(\sigma_i^k)^2$ are summable, we have that $\sum_{i=1}^m\|x^{k+1}_i - \bar x^{k+1} \|^2$ satisfies all conditions in  Lemma \ref{Lemma-polyak_2}
in the appendix with $v^k=\sum_{i=1}^m\|x^{k+1}_i - \bar x^{k+1} \|^2$, $\alpha^k=0$, $q^k=1-\rho$, and $p^k=\frac{m^2d^2\max\limits_{i}\left((\bar\lambda_i^k)^2+(\sigma_i^k)^2\right)}{1-\rho}G^2$. Therefore, under the given conditions in the statement, we have $\sum_{i=1}^m\|x^{k+1}_i - \bar x^{k+1} \|^2$ converging to 0 almost surely, and hence the result in (\ref{eq:consensus}).

We next proceed to prove (\ref{eq:gradient_summable1}) and (\ref{eq:graident_summable2}).

}

From the Lipschitz gradient condition  in Assumption
\ref{Assumption:gradient}, we have
\[
F(y)\leq F(x)+ \langle \nabla F(x),(y-x)\rangle +\frac{L\|y-x\|^2}{2}
\]
By plugging $y=\bar{x}^{k+1}$ and $x=\bar{x}^{k}$ into the above
inequality, we obtain the following relationship based on
(\ref{eq:x_bar_dynamics}):
\begin{equation}\label{eq:f_X_k+1_secondstep}
\begin{aligned}
\mathbb{E}\left[F(\bar{x}^{k+1})\right]&\leq
\mathbb{E}\left[F(\bar{x}^k)\right]-\mathbb{E}\left[\left\langle
\nabla F(\bar{x}^k),
\frac{1}{m}\sum_{i=1}^m \Lambda_i^k g_i^k\right\rangle \right]\\
&\qquad+ \frac{L}{2m^2}\mathbb{E}\left[ \left\|\sum_{i=1}^m
\Lambda_i^k g_i^k \right\|^2\right]
\end{aligned}
\end{equation}

Next we analyze the   last two terms on the right hand side of
(\ref{eq:f_X_k+1_secondstep}).
Defining $\Lambda^k=[\Lambda_1^k,\cdots,\Lambda_m^k]^T\in\mathbb{R}^{d\times md}$, we have
\[
\left\| \sum_{i=1}^m
\Lambda_i^kg_i^k\right\|^2=\|\Lambda^k g^k\|^2\leq \|\Lambda^k\|^2 \|g^k\|^2
\]

Using the  relationship $\| \Lambda^k\|^2\leq \| \Lambda^k\|_F^2=\sum_{j=1}^{m}\sum_{p=1}^{d}(\lambda_{jp}^k)^2$ and the independence of $\Lambda_i^k$ and $g_i^k$,
 we have
\begin{equation}\label{eq:first_term}
\begin{aligned}
&\frac{L}{2m^2}\mathbb{E}\left[\left\| \sum_{i=1}^m
\Lambda_i^kg_i^k\right\|^2\right] =\frac{L}{2m^2}  \sum_{i=1}^m md G^2((\bar\lambda_i^k)^2+(\sigma_i^k)^2)
\end{aligned}
\end{equation}

The second last term on the right hand side of
(\ref{eq:f_X_k+1_secondstep}) can be rewritten as
\begin{equation}\label{eq:inner_purduct}
\begin{aligned}
&\mathbb{E}\left[\left\langle  \nabla  F(\bar{x}^k),
\frac{1}{m}\sum_{i=1}^m
\Lambda_i^k g_i^k \right\rangle\right]\\
&={ \mathbb{E}\left[\left\langle  \nabla
F(\bar{x}^k),\frac{1}{m}\sum_{i=1}^m\mathbb{E}\left[
  \Lambda_i^k g_i^k|\mathcal{F}^k\right]\right\rangle\right]} \\
&=\mathbb{E}\left[\left\langle  \nabla
F(\bar{x}^k),\frac{1}{m}\sum_{i=1}^m
  \bar\lambda_i^k\nabla f_i(x_i^k) \right\rangle\right]
%
%
\end{aligned}
\end{equation}
Note that in the first equality, { the expectation in the
inner product is a conditional expectation taken with respect to
$\mathcal{F}^k$, the sigma algebra generated by randomness
$\{\Lambda_1^k,\xi_1^k,\cdots,\Lambda_m^k,\xi_m^k\}$ at iteration
$k$}.  The expectation outside the inner product is the full
expectation taken over the sigma algebra generated by randomness in
stepsizes and gradients in all iterations up to $k$.

 Given that  for any $1\leq p \leq m$, we have
\[
\begin{aligned}
 \sum_{i=1}^m\bar\lambda_i^k\nabla f_i(x_i^k)    = \sum_{i=1}^m\bar\lambda_p^k\nabla f_i(x_i^k)
+ \sum_{i=1}^m(\bar\lambda_i^k-\bar\lambda_p^k)\nabla f_i(x_i^k)
\end{aligned}
\]
we can further rewrite (\ref{eq:inner_purduct}) as
%
\[
\begin{aligned}
&\mathbb{E}\left[\left\langle  \nabla  F(\bar{x}^k),
\frac{1}{m}\sum_{i=1}^m \Lambda_i^k g_i^k \right\rangle\right]
\\
&=\mathbb{E}\left[\left\langle  \nabla  F(\bar{x}^k),
\frac{\bar\lambda_p^k }{m}\sum_{i=1}^m \nabla f_i(x_i^k)
\right\rangle\right] \\
&\qquad + \mathbb{E}\left[\left\langle  \nabla F(\bar{x}^k),
 \frac{1}{m}\sum_{i=1}^m
(\bar\lambda_i^k-\bar\lambda_p^k) \nabla f_i(x_i^k) \right\rangle\right]\\
&=\bar\lambda_p^k \mathbb{E}\left[\left\langle  \nabla F(\bar{x}^k),
\frac{ 1}{m} \sum_{i=1}^m \nabla f_i(x_i^k)\right\rangle\right]\\
&\qquad + \mathbb{E}\left[\left\langle \nabla F(\bar{x}^k),
 \frac{1}{m}\sum_{i=1}^m
(\bar\lambda_i^k-\bar\lambda_p^k) \nabla f_i(x_i^k)
\right\rangle\right]
\end{aligned}
\]

Applying the relationship  $2\langle
X,Y\rangle=\|X\|^2+\|Y\|^2-\|X-Y\|^2$ to the first inner product
term on the right hand side of the above inequality leads to

\begin{equation}\label{eq:nonconvex_revision}
\begin{aligned}
&\mathbb{E}\left[\left\langle  \nabla  F(\bar{x}^k),
\frac{1}{m}\sum_{i=1}^m \Lambda_i^k g_i^k \right\rangle\right]\\
&=\frac{\bar\lambda_p^k}{2}\mathbb{E}\left[ \left\|\nabla
F(\bar{x}^k)\right\|^2 +
 \left\|\frac{\sum_{i=1}^m \nabla
f_i(x_i^k)  }{m}\right\|^2 \right.\\
&\qquad\qquad\qquad \left.- \left\| \nabla F(\bar{x}^k) -\sum_{i=1}^m
\frac{\nabla
f_i(x_i^k)}{m}\right\|^2  \right] \\
&\quad +\mathbb{E}\left[\left\langle \nabla  F(\bar{x}^k), \frac{1}
{m}\sum_{i=1}^m(\bar\lambda_i^k-\bar\lambda_p^k)\nabla
f_i(x_i^k)\right\rangle\right]\\
&\geq  \frac{\bar\lambda_p^k}{2}\mathbb{E}\left[ \left\|\nabla
F(\bar{x}^k)\right\|^2 +
 \left\|\frac{\sum_{i=1}^m \nabla
f_i(x_i^k)}{m}  \right\|^2\right.\\
&\qquad\qquad\qquad \left.- \frac{L^2}{m^2}\sum_{i=1}^m\|\bar{x}^k- x_i^k\|^2  \right] \\
&\quad +\mathbb{E}\left[\left\langle \nabla  F(\bar{x}^k), \frac{1}
{m}\sum_{i=1}^m(\bar\lambda_i^k-\bar\lambda_p^k)\nabla
f_i(x_i^k)\right\rangle\right]
\end{aligned}
\end{equation}
where we used the Lipschitz assumption on the gradient in Assumption
\ref{Assumption:gradient}.

{ According to the Cauchy-Schwarz inequality, we have
$\left\langle u, v\right\rangle\geq -\|u\|\|v\|$, which leads to the
following relationship for the last term in
(\ref{eq:nonconvex_revision}):}
\begin{equation}\label{eq:last_term_inner_product}
{ \begin{aligned} &\mathbb{E}\left[\left\langle \nabla
F(\bar{x}^k), \frac{1}
{m}\sum_{i=1}^m(\bar\lambda_i^k-\bar\lambda_p^k)\nabla
f_i(x_i^k)\right\rangle\right]\\
& \geq -\mathbb{E}\left[ \left\|\nabla  F(\bar{x}^k)\right\|
\left\|\frac{1}
{m}\sum_{i=1}^m(\bar\lambda_i^k-\bar\lambda_p^k)\nabla
f_i(x_i^k)\right\|\right]\\
& \geq -\mathbb{E}\left[ \left\|\nabla  F(\bar{x}^k)\right\|  \frac{1}
{m}\sum_{i=1}^m|\bar\lambda_i^k-\bar\lambda_p^k|\left\|\nabla
f_i(x_i^k)\right\|\right] \\
&\geq -\mathbb{E}\left[ \frac{G^2}
{m}\sum_{i=1}^m|\bar\lambda_i^k-\bar\lambda_p^k|\right]=- \frac{G^2}
{m}\sum_{i=1}^m|\bar\lambda_i^k-\bar\lambda_p^k |
\end{aligned}
}
\end{equation}
where in the last inequality we used the boundedness of gradients
$\nabla F(\bar{x}^k)$ and $\nabla f_i(x_i^k)$ by $G$.

Combining (\ref{eq:nonconvex_revision}) and
(\ref{eq:last_term_inner_product})  leads to
\begin{equation}\label{eq:inner_product_final}
\begin{aligned}
&\mathbb{E}\left[\left\langle  \nabla  F(\bar{x}^k),
\frac{1}{m}\sum_{i=1}^m
\Lambda_i^k g_i^k\right\rangle\right]\\
&\geq  \frac{\bar\lambda_p^k}{2}\mathbb{E}\left[ \left\|\nabla
F(\bar{x}^k)\right\|^2\right] +
 \frac{\bar\lambda_p^k }{2}\mathbb{E}\left[\left\|\frac{\sum_{i=1}^m\nabla
f_i(x_i^k)}{m} \right\|^2\right]\\
&\quad -\frac{\bar\lambda_p^kL^2}{2m^2}\sum_{i=1}^m \mathbb{E}\left[ \|\bar{x}^k- x_i^k\|^2 \right]    -\frac{G^2}{m}\sum_{i=1}^m |\bar\lambda_i^k-\bar\lambda_p^k|
\end{aligned}
\end{equation}

Plugging (\ref{eq:first_term}) and (\ref{eq:inner_product_final})
into (\ref{eq:f_X_k+1_secondstep}) leads to

\begin{equation}\label{eq:f_X_k+1_thirdstep}
\begin{aligned}
&\mathbb{E}\left[F(\bar{x}^{k+1})\right]\leq
\mathbb{E}\left[F(\bar{x}^k)\right]\\
& -\frac{\bar\lambda_p^k}{2}\mathbb{E}\left[ \left\|\nabla
F(\bar{x}^k)\right\|^2\right] -
 \frac{ \bar\lambda_p^k}{2}  \mathbb{E}\left[\left\|\frac{\sum_{i=1}^m\nabla
f_i(x_i^k) }{m}\right\|^2\right]\\
&+\frac{\bar\lambda_p^k L^2}{2m}\sum_{i=1}^m\mathbb{E}\left[
\|\bar{x}^k-x_i^k \|^2\right]
+\frac{G^2}{m}\sum_{i=1}^m |\bar\lambda_i^k-\bar\lambda_p^k|\\
&+\frac{L\sum_{i=1}^mmd G^2((\bar\lambda_i^k)^2+(\sigma_i^k)^2)}{2m^2}
\end{aligned}
\end{equation}
or
\begin{equation}\label{eq:f_X_k+1_thirdstep}
\begin{aligned}
&\bar\lambda_p^k\mathbb{E}\left[ \left\|\nabla
F(\bar{x}^k)\right\|^2\right] +
 \bar\lambda_p^k  \mathbb{E}\left[\left\|\frac{\sum_{i=1}^m\nabla
f_i(x_i^k) }{m}\right\|^2\right]\\
&\leq 2\mathbb{E}\left[F(\bar{x}^k)\right]-2\mathbb{E}\left[F(\bar{x}^{k+1})\right]\\
& +\frac{\bar\lambda_p^k L^2}{m}\sum_{i=1}^m\mathbb{E}\left[
\|\bar{x}^k-x_i^k \|^2\right]
+\frac{2G^2}{m}\sum_{i=1}^m |\bar\lambda_i^k-\bar\lambda_p^k|\\
& +\frac{L\sum_{i=1}^m md G^2((\bar\lambda_i^k)^2+(\sigma_i^k)^2)}{2m^2}
\end{aligned}
\end{equation}

Adding (\ref{eq:f_X_k+1_thirdstep}) from $k=0$ to $k=t$ yields

\begin{equation}
\begin{aligned}
&\sum_{k=0}^t\bar\lambda_p^k\mathbb{E}\left[ \left\|\nabla
F(\bar{x}^k)\right\|^2\right] +
 \sum_{k=0}^t \bar\lambda_p^k \mathbb{E}\left[\left\|\frac{\sum_{i=1}^m\nabla
f_i(x_i^k) }{m}\right\|^2\right]\\
&\leq 2\mathbb{E}\left[F(\bar{x}^0)\right]-2\mathbb{E}\left[F(\bar{x}^{t+1})\right]\\
& +\sum_{k=0}^t\frac{\bar\lambda_p^k
L^2}{m}\sum_{i=1}^m\mathbb{E}\left[ \|\bar{x}^k-x_i^k \|^2\right]
+\sum_{k=0}^t\frac{2G^2}{m}\sum_{i=1}^m |\bar\lambda_i^k-\bar\lambda_p^k|\\
& +\sum_{k=0}^t\frac{L\sum_{i=1}^m md G^2((\bar\lambda_i^k)^2+(\sigma_i^k)^2)}{m^2}
\end{aligned}
\end{equation}
or
\begin{equation}\label{eq:adding_from_0_to_t}
\begin{aligned}
&\frac{\sum_{k=0}^t\bar\lambda_p^k\mathbb{E}\left[ \left\|\nabla
F(\bar{x}^k)\right\|^2\right]}{\sum_{k=0}^t\bar\lambda_p^k} +
 \frac{\sum_{k=0}^t \bar\lambda_p^k \mathbb{E}\left[\left\|\frac{\sum_{i=1}^m\nabla
f_i(x_i^k) }{m}\right\|^2\right]}{\sum_{k=0}^t\bar\lambda_p^k}\\
&\leq\frac{2\mathbb{E}\left[F(\bar{x}^0)\right]-2\mathbb{E}\left[F(\bar{x}^{t+1})\right]}{\sum_{k=0}^t\bar\lambda_p^k}\\
&+\frac{\sum_{k=0}^t\frac{\bar\lambda_p^k
L^2}{m}\sum_{i=1}^m\mathbb{E}\left[ \|\bar{x}^k-x_i^k
\|^2\right]}{\sum_{k=0}^t\bar\lambda_p^k}
\\
&+\frac{\sum_{k=0}^t\frac{2G^2}{m}\sum_{i=1}^m |\bar\lambda_i^k-\bar\lambda_p^k|}{\sum_{k=0}^t\bar\lambda_p^k}\\
&+\frac{\sum_{k=0}^t\frac{L}{m}\sum_{i=1}^md G^2((\bar\lambda_i^k)^2+(\sigma_i^k)^2)}{\sum_{k=0}^t\bar\lambda_p^k}
\end{aligned}
\end{equation}

Invoking the proven result in (\ref{eq:consensus}), we know that the second
term on the right hand side of (\ref{eq:adding_from_0_to_t}) will
converge to zero as $t\rightarrow \infty$. Using the conditions on
the expectation and variance of stepsizes, it can be obtained that
the third and fourth terms on the right hand side of
(\ref{eq:adding_from_0_to_t}) will also converge to zero as
$t\rightarrow \infty$. Therefore, we   have the results in Theorem
\ref{theorem_non_convex}.

\end{proof}

 {  \begin{Remark 1}
  Eqn. (\ref{eq:adding_from_0_to_t})
indicates that the convergence speed of  $\mathbb{E}\left[\left\|\nabla F(\bar{x}^k)\right\|^2\right]$ and $
\mathbb{E}\left[\left\|\frac{\sum_{i=1}^m\nabla f_i(x_i^k)
}{m}\right\|^2\right]$   to zero is determined by the convergence speed of $\mathbb{E}\left[ \|\bar{x}^k-x_i^k
\|^2\right]$, $|\bar\lambda_i^k-\bar\lambda_p^k|$, and $(\bar\lambda_i^k)^2+(\sigma_i^k)^2$ to zero. From (\ref{eq:x^{k+1}-barx_norm3}), one obtains that the convergence speed of $\|\bar{x}^k-x_i^k
\|^2$ to zero is no slower than $\mathcal{O}(\frac{1}{k})$. According to the assumption, i.e., all $|\bar\lambda_i^k-\bar\lambda_p^k|$, $(\bar\lambda_i^k)^2$, and $(\sigma_i^k)^2$ are summable, one has that they all decrease  to zero no slower than  $\mathcal{O}(\frac{1}{k})$. Therefore, we have that $\mathbb{E}\left[\left\|\nabla F(\bar{x}^k)\right\|^2\right]$ and $
\mathbb{E}\left[\left\|\frac{\sum_{i=1}^m\nabla f_i(x_i^k)
}{m}\right\|^2\right]$ converge to zero no slower than $\mathcal{O}(\frac{1}{k})$, i.e., $\mathbb{E}\left[\left\|\nabla F(\bar{x}^k)\right\|\right]$ and $
\mathbb{E}\left[\left\|\frac{\sum_{i=1}^m\nabla f_i(x_i^k)
}{m}\right\|\right]$ converge to zero no slower than $\mathcal{O}(\frac{1}{k^{0.5}})$.
\end{Remark 1}}

Although the convergence of  weighted sum of gradients  in
(\ref{eq:gradient_summable1}) and (\ref{eq:graident_summable2}) are
the most commonly used form of convergence for stochastic gradient
descent under diminishing stepsizes, it only implies that the limit
inferior of gradients will be zero \cite{bottou2018optimization},
i.e., $ \underline\lim_{k\rightarrow \infty}
\mathbb{E}\left[\left\|\nabla F(\bar{x}^k)\right\|^2\right]=0$ and $
\underline\lim_{k\rightarrow \infty}
\mathbb{E}\left[\left\|\frac{\sum_{i=1}^m\nabla f_i(x_i^k)
}{m}\right\|^2\right]=0$,  which means that the mathematical
expectation of gradients cannot stay bounded away from zero.
 In fact, by adding an additional assumption on the Lipschitz continuity of the Hessian $H(f_i(\cdot))$ of function $f_i(\cdot)$ , we can guarantee the convergence of the expectation of gradients
 to zero, which is stated below:
\begin{Theorem 1}\label{Theorem:non_convex_exactconvergence}
Under the conditions of Theorem \ref{theorem_non_convex}, if we
further have {
\begin{equation}\label{eq:additional_assumption}
\|H(f_i(x))- H(f_i(y))\|\leq \nu\|x-y\|, \quad \forall 1\leq i\leq m
\end{equation}}
for a $\nu>0$ under any $x$ and $y$, then we can obtain
\[
\lim\limits_{k\rightarrow \infty} \mathbb{E}\left[\left\|\nabla
F(\bar{x}^k)\right\|^2\right]=0
\]
and
\[
\lim\limits_{k\rightarrow \infty}
\mathbb{E}\left[\left\|\frac{\sum_{i=1}^m\nabla f_i(x_i^k)
}{m}\right\|^2\right]=0
\]
\end{Theorem 1}
\begin{proof}
The results can be obtained following the same line of derivation of Theorem 5 in \cite{george2019distributed} and hence we omit the proof here.
\end{proof}

\begin{Remark 1}
The proposed inherently privacy-preserving decentralized SGD
algorithm can maintain the accuracy of optimization results while
enabling privacy protection. This is in distinct difference from
differential-privacy based privacy solutions that have to trade
optimization accuracy for privacy.
\end{Remark 1}

\section{Privacy analysis}
In this section, we analyze the inherent privacy embedded in our
proposed algorithm against adversaries which try to infer the
gradient of participating agents based on received information. Note
that both $b_{ij}^k$ and $\Lambda_j^k$ are determined by agent $j$
randomly and hence can provide privacy protection for the gradient
$g_j^k$ of agent $j$. Also note that when the gradient is a vector  with dimension $d>1$, $\Lambda_j^k$
is a diagonal matrix  $\Lambda_j^k={\rm
diag}\{\lambda_{j1}^k,\lambda_{j2}^k,\cdots,\lambda_{jd}^k\}$ with
all elements  independently chosen  by agent $j$ (under the same
mathematical expectation and variance, i.e.,
$\mathbb{E}\Lambda_j^k=\bar\lambda_j^k I_d$ and $ {\rm var} \left[\Lambda_j^k\right]=(\sigma_j^k)^2 I_d$) to obfuscate each
element of $g_j^k$ independently. However, $b_{ij}^k$ would still be
a scalar (although random) when $d>1$, which limits the privacy it can provide
in obfuscating  $g_j^k$. {(For example, the ratio
between different entries of $g_j^k$ will not be covered by
$b_{ij}^k$ if a scalar $b_{ij}^k$
 were used alone to obfuscate $g_j^k$.)} So here we only consider the
privacy enabled by the stepsize $\Lambda_j^k$, which is not affected
by the dimension of $g_j^k$. { Note that $b_{ij}^k$
still contributes to privacy protection since it increases the
number of unknown parameters in the system and hence enhances
resilience to potential side information attacks. For example, the
usage of $b_{ij}^k$ can  avoid an adversary from inferring $g_j^k$
even after the adversary obtains (for whatever reason)  the values
of $\Lambda_j^k$ and $x_j^k$.}

 { Because each element of $g_j^k$ is independently obscured by a corresponding
  element  of $\Lambda_j^k$, we only need to consider the
scalar case where $\Lambda_j^k$ becomes a scalar $\lambda_j^k$ and provides protection for a scalar gradient $g_j^k$.
This is because when $d>1$, if  we can quantify the privacy that the
$i$th element of $\Lambda_j^k$ (i.e., $\lambda_{ji}^k$) enables for
the $i$th entry of vector $g_j^k$ (i.e., $g_{ji}^k$),   we can also
quantify  the privacy that other entries  of $\Lambda_j^k$ (i.e.,
$\lambda_{jq}^k$ for $1\leq q\leq d,\,q\neq i$) provide  for other
entries of vector $g_j^k$ (i.e., $g_{jq}^k$ for $1\leq q\leq
d,\,q\neq i$). According to the basic relationship
(\ref{eq:conditional_entropy_basic}) in information theory, if we
can quantify the conditional differential entropy
$h(g_j^k|\lambda_j^kg_j^k)$, then we can obtain an adversary's best
estimation accuracy and hence evaluate the privacy enabled by the
random stepsize $\lambda_j^k$. Clearly, when $\lambda_j^k$ is
deterministic and homogeneous across all agents (and hence publicly
known, which is the case in most existing algorithms that do not
consider privacy protection), the conditional differential entropy
$h(g_j^k|\lambda_j^kg_j^k)$ will be zero because an adversary can
uniquely determine $g_j^k$ from accessed $\lambda_j^kg_j^k$. Next,
we quantify the privacy enabled by the random stepsize $\lambda_j^k$
in our algorithm.}

Without loss of generality, we consider the case where $g_j^k$ is
uniformly distributed in the interval $[-\kappa,\,\kappa]$ where
$\kappa$ is a positive scalar. {  Note   we assume that
this range information of gradients (usually called side
information) is available to adversaries in evaluating the
protection performance against their inference attacks.} The same
argument can be easily adapted to cases where $g_j^k$ has other
probability distributions. To obscure the gradient information, we
use a random $\lambda_j^k$ uniformly distributed in
$[0,\,2\bar\lambda_j^k]$, which can be verified to have expectation
$\mathbb{E}\left[\lambda_j^k\right]=\bar\lambda_j^k$ and standard
deviation $\sigma(\lambda_j^k)=\frac{\sqrt{3}\bar\lambda_j^k}{3}$.
{ Note that because $\lambda_j^k$  can be zero when
uniformly distributed in $[0,\,2\bar\lambda_j^k]$, an adversary
cannot uniquely determine that $g_j^k$ is zero even if it (for
whatever reason) knows that the product $\lambda_j^k g_j^k$ is zero.
}  For the sake of simplicity, we assume $2\bar\lambda_j^k\leq
\kappa$. Then we can obtain the following result about the
adversary's best estimation accuracy:

\begin{Theorem 2}\label{theorem:estimation_error}
Under our decentralized stochastic gradient descent algorithm, if an
eavesdropping or honest-but-curious adversary tries to estimate  the
gradient $ g_j^k$ of an agent based on received information, its
estimation error will be no less than
$\frac{e^{2\vartheta(\bar\lambda_j^k,\kappa)}}{2\pi e}$, where
\begin{equation}\label{eq:h(lambda_kappa)}
\vartheta(\bar\lambda_j^k,\kappa)=
\log(4\bar\lambda_j^k\kappa^2)-1-c(\bar\lambda_j^k,\kappa)
\end{equation}
with $c(\bar\lambda_j^k,\kappa)$ given by
\begin{equation}\label{eq:entropy_multiplication}
\begin{aligned}
&c(\bar\lambda_j^k,\kappa)=
-2\int_{0}^{2\bar\lambda_j^k\kappa}\frac{\log(\frac{2\bar\lambda_j^k\kappa}{x})}{4\bar\lambda_j^k\kappa}\log\left(\frac{\log(\frac{2\bar\lambda_j^k\kappa}{x})}{4\bar\lambda_j^k\kappa}\right)dx
\end{aligned}
\end{equation}

\end{Theorem 2}
\begin{proof}
To prove that the estimation error is not less than
$\frac{e^{2\vartheta(\bar\lambda_j^k,\kappa)}}{2\pi e}$, we show that the
conditional differential entropy $h(g_j^k|\lambda_j^k g_j^k)$ is no
less than $\vartheta(\lambda_j^k,\kappa)$ in (\ref{eq:h(lambda_kappa)}).
According to information theory,
\[
h(g_j^k|\lambda_j^k g_j^k)=h(g_j^k, \lambda_j^kg_j^k)-h( \lambda_j^k
g_j^k)
\]
where $h(g_j^k, \lambda_j^k g_j^k)$ denotes the joint differential
entropy of $g_j^k$ and the product variable $ \lambda_j^k g_j^k$,
and $h(\lambda_j^k g_j^k)$ represents the differential entropy of
$\lambda_j^k g_j^k$. Next we first calculate the joint differential
entropy of $g_j^k$ and the product variable $\lambda_j^k g_j^k$.
According to information theory, we have
\begin{equation}\label{eq:joint_entropy}
\begin{aligned}
&h(g_j^k,\lambda_j^k g_j^k)\\
&=-\int_{g_j^k}\int_{\lambda_j^k
g_j^k}p(g_j^k,\lambda_j^kg_j^k)\log(p(g_j^k,\lambda_j^kg_j^k))d\lambda_j^kg_j^kdg_j^k
\end{aligned}
\end{equation}
where $p(\cdot)$ represents the probability density function.

Making use of the fact that $\lambda_{j}^k$  and $g_j^k$ are
stochastically independent random variables with uniform
distributions, we have
\[
\begin{aligned}
p(g_j^k,\lambda_j^kg_j^k)= p(\lambda_j^kg_j^k|g_{j}^k)p(g_j^k)
 =\frac{1}{2\bar\lambda_j^k|g_j^k|}\frac{1}{2\kappa}=\frac{1}{4\bar\lambda_j^k\kappa |g_j^k|}
\end{aligned}
\]
Substituting this relationship into    (\ref{eq:joint_entropy})
leads to
\[
\begin{aligned}
&h(g_j^k, \lambda_j^k g_j^k)\\
&=-\int_{g_j^k}\int_{\lambda_j^k
g_j^k}\frac{1}{4\bar\lambda_j^k\kappa
|g_j^k|}\log(\frac{1}{4\bar\lambda_j^k\kappa
|g_j^k|})d\lambda_j^kg_j^kdg_j^k\\
&=-\int_{g_j^k}\int_{\lambda_j^k}\frac{1}{4\bar\lambda_j^k\kappa
|g_j^k|}g_j^k\log(\frac{1}{4\bar\lambda_j^k\kappa
|g_j^k|})d\lambda_j^kdg_j^k\\
&=-\int_{g_j^k}2\bar\lambda_j^k\frac{1}{4\bar\lambda_j^k\kappa
|g_j^k|}g_j^k\log(\frac{1}{4\bar\lambda_j^k\kappa |g_j^k|})dg_j^k\\
&=-\int_{g_j^k}
\frac{g_j^k}{2\kappa|g_j^k|}\log(\frac{1}{4\bar\lambda_j^k\kappa
|g_j^k|})dg_j^k\\
&=-\int_{-\kappa}^0
\frac{-1}{2\kappa}\log(\frac{1}{-4\bar\lambda_j^k\kappa
g_j^k})dg_j^k-\int_{0}^{\kappa}
\frac{1}{2\kappa}\log(\frac{1}{4\bar\lambda_j^k\kappa g_j^k})dg_j^k\\
&=\int_{0}^{\kappa} \frac{1}{\kappa}\log(4\bar\lambda_j^k\kappa
g_j^k)dg_j^k\\
&=\log(4\bar\lambda_j^k\kappa^2)-1
\end{aligned}
\]
To calculate the differential entropy of the product variable
$\lambda_j^kg_j^k$, we first determine its probability distribution
$p(\lambda_j^kg_j^k)$. It can be seen that $\lambda_j^kg_j^k$
resides in the interval
$(-2\kappa\bar\lambda_j^k,\,2\kappa\bar\lambda_j^k)$. We first
determine the probability distribution of $\lambda_j^kg_j^k$ on the
interval $(0,2\kappa\bar\lambda_i^k]$. The cumulative probability of
$\lambda_j^kg_j^k$ on any given interval $(0,x]$ for $x\leq
2\kappa\bar\lambda_j^k$ is given by
\begin{equation}\label{eq:prob}
\begin{aligned}
P(0\leq 2\lambda_j^k g_j^k \leq x)&=\int_0^{\kappa}
P(0\leq\lambda_j^k g_j^k \leq x | g_j^k)p(g_j^k)d g_j^k\\
&=\int_0^{\kappa} P(0\leq \lambda_j^k  \leq
\frac{x}{g_j^k})p(g_j^k)d g_j^k
\end{aligned}
\end{equation}
where $P(\cdot)$ denotes the cumulative probability.

Because we have $P(0\leq \lambda_j^k  \leq \frac{x}{g_j^k})=1$ when
$g_j^k\leq \frac{x}{2\bar\lambda_j^k}$ and $P(0\leq \lambda_j^k \leq
\frac{x}{g_j^k})=\frac{x}{g_j^k2\bar\lambda_j^k}$ when $g_j^k\geq
\frac{x}{2\bar\lambda_j^k}$, we can rewrite (\ref{eq:prob}) as
\begin{equation}\label{eq:prob1}
\begin{aligned}
&P(0\leq 2\lambda_j^k g_j^k \leq x)\\
&\qquad =\int_0^{\frac{x}{2\bar\lambda_j^k}} 1p(g_j^k)d g_j^k+
\int_{\frac{x}{2\bar\lambda_j^k}}^{\kappa}
\frac{x}{g_j^k2\bar\lambda_j^k}p(g_j^k)d g_j^k\\
&\qquad =\int_0^{\frac{x}{2\bar\lambda_j^k}} \frac{1}{2\kappa}d
g_j^k+ \int_{\frac{x}{2\bar\lambda_j^k}}^{\kappa}
\frac{x}{g_j^k2\bar\lambda_j^k}\frac{1}{2\kappa}d g_j^k\\
&\qquad
=\frac{x}{4\kappa\bar\lambda_j^k}+\frac{x}{4\kappa\bar\lambda_j^k}(\log\kappa-\log(\frac{x}{2\bar\lambda_j^k}))
\end{aligned}
\end{equation}
Then for $0\leq \lambda_j^kg_j^k\leq 2\lambda_j^k\kappa$, the
probability density function can be obtained as
\[
p(\lambda_j^kg_j^k=x)=\frac{d}{dx}P(0\leq 2\lambda_j^k g_j^k \leq
x)=\frac{\log(\frac{2\bar\lambda_j^k\kappa}{x})}{4\bar\lambda_j^k\kappa}
\]

Similarly, we can get the probability density function of
$\lambda_j^kg_j^k$ in the interval $-2\lambda_j^k\kappa\leq
\lambda_j^kg_j^k\leq 0$ as
\[
p(\lambda_j^kg_j^k=x)=
\frac{\log(\frac{2\bar\lambda_j^k\kappa}{-x})}{4\bar\lambda_j^k\kappa}
\]
Knowing the probability density function of $\lambda_j^kg_j^k$
enables us to calculate its differential entropy, which is exactly
  (\ref{eq:entropy_multiplication}).

\end{proof}

\begin{Remark 1}
According to the convergence results in Theorem
\ref{Theorem:Theorem_1} - Theorem
\ref{Theorem:non_convex_exactconvergence}, the amplitude of the
expected stepsize $\bar\lambda_j^k$ has to decrease to $0$ to
guarantee the claimed convergence results. {  Different from additive-noise  based privacy approaches which will lose   protection completely when the noise decreases to zero, the stochastic  stepsize in our approach obfuscates information in a multiplicative manner and hence still provides  protection even when it converges to zero. More specifically, when $\bar\lambda_j^k$ converges to zero, the conditional differential entropy will still be non-zero (due to the uncertainty in gradient $g_j^k$). In this case, the multiplicative relationship between $\lambda_j^k$ and $g_j^k$ makes  an adversary's observation $\lambda_j^k g_j^k$ be always equal to $0$ irrespective of the value of $g_j^k$, and hence avoids the adversary from inferring the value of $g_j^k$. In fact, using the Matlab numerical integration function ``integral", we can obtain that under $\kappa=5$, when $\bar\lambda_j^k$
is set  to zero, the expression $\vartheta(\bar\lambda_j^k,\kappa)$ in (\ref{eq:h(lambda_kappa)}) (a logarithm function minus a constant and a  definite integral)  is equal  to
$1.0322$, meaning that the  conditional differential entropy is no less than $1.0322$. Therefore,  according to the relationship in (\ref{eq:conditional_entropy_basic}), by taking exponential and dividing by $2\pi e$, we can obtain that the adversary's best expected squared estimation
error is   no less than $0.4614$ after $\bar\lambda_j^k$ converges to zero.}
\end{Remark 1}

\begin{Remark 1}
Following the same line of  derivation, we can also obtain an
adversary's best estimation error when the  gradients follow other
probability distributions.
\end{Remark 1}
\begin{Remark 1}
{  It can be verified that under the considered uniform distribution of $\lambda_j^k$, a larger $\bar{\lambda}_j^k$ (and hence a larger variance $\frac{(\bar\lambda_j^k)^2}{3}$ of the stepsize) means a larger conditional differential entropy and hence better privacy protection. But as long as $\bar{\lambda}_j^k$ satisfies the non-summable but square summable condition  in (\ref{eq:non_summable}) and the summable heterogeneity condition in (\ref{eq:heterogeneity_condition}), convergence can always be achieved.} Moreover, the proposed algorithm can guarantee the privacy of every
participating agent even when an adversary can have access to all
shared messages in the network. This is in distinct difference from
existing accuracy-friendly privacy-preserving solutions  (in, e.g.,
\cite{yan2012distributed,lou2017privacy,gade2018privacy} for
decentralized deterministic convex optimization) that have to
restrict an adversary from having full access to   messages on all
channels to enable privacy protection.
\end{Remark 1}

{ \begin{Remark 1} It is worth noting that our algorithm
can also provide privacy protection for the variable $x$. This is
because the actual variable shared by agent $j$  is
$v_{ij}^k=w_{ij}x_j^k-b_{ij}^k\Lambda_j^kg_j^k$. Since $x_j^k$,
$b_{ij}^k$, and $\Lambda_j^k$ are private to agent $j$ and unknown
to adversaries, the value of $x_j^k$ is not disclosed from shared
$v_{ij}^k$. However, note that since all states $x_j^k$ will finally
reach consensus and converge to the same value, the final $x_j^k$
will be disclosed when $k$ tends to infinity. Therefore, our
algorithm can avoid the intermediate state value  $x_j^k$ of every
agent $j$ from disclosure before the algorithm converges (i.e.,
before all states $x_j^k$ reach consensus).
\end{Remark 1}}
%

\section{Numerical experiments}
In this section, we evaluate the performance of our algorithm using
numerical experiments. We will consider both the
convex-objective-function case and the non-convex-objective-function
case.
\subsection{Convex case}
For the case of convex objective functions, we consider a canonical
decentralized estimation problem where a sensor network of $m$
sensors collectively estimate an unknown parameter
$\theta\in\mathbb{R}^d$, which can be formulated as an empirical
risk minimization problem. More specifically, we assume that each
sensor $i$ has $n_i$ noisy measurements of the parameter
$z_{ij}=M_i\theta+w_{ij}$ for $j=\{1,2,\cdots,n_i\}$ where
$M_i\in\mathbb{R}^{s\times d}$ is the measurement matrix of agent
$i$ and $w_{ij}$ is measurement noise associated with measurement
$z_{ij}$. Then the estimation of the parameter $\theta$ can be
solved using empirical risk minimization problem formulated as
(\ref{eq:optimization_formulation1}), with each $f_i(\theta)$ given
as
\[
f_i(\theta)=\frac{1}{n_i}\sum_{j=1}^{n_i}\|z_{ij}-M_i\theta\|^2+r_i\|\theta\|^2
\]
where $r_i$ is a non-negative regularization parameter.

\begin{figure}
    \begin{center}
        \includegraphics[width=0.35\textwidth]{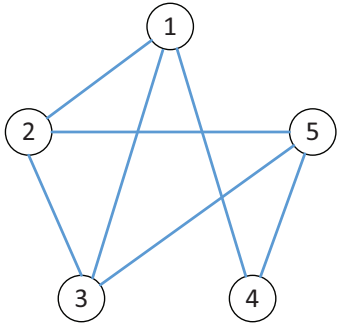}
    \end{center}
    \caption{The interaction topology of the network.}
    \label{fig:topology}
\end{figure}

\begin{figure}
    \begin{center}
        \includegraphics[width=0.5\textwidth]{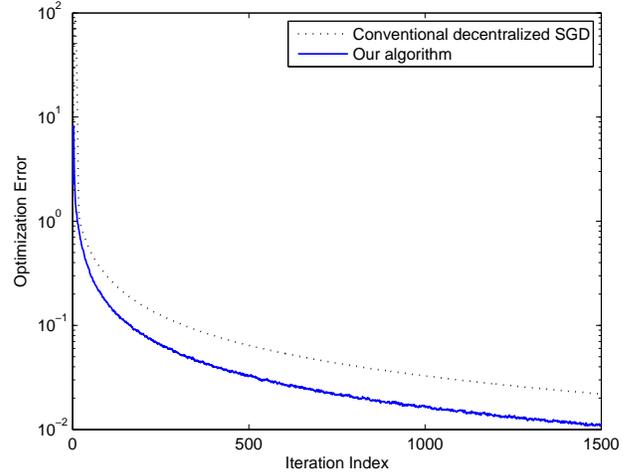}
    \end{center}
    \caption{Comparison of convergence performance between our algorithm and the conventional decentralized SGD in \cite{lian2017can}.}
    \label{fig:comp_convex}
\end{figure}

We assume that the network  consists  of five agents interacting on
a graph depicted in Fig. \ref{fig:topology}. The parameter $s$ was
set to three and the parameter $d$ was set to two without loss of
generality. $n_i$ was set to 100 for all $i$. $w_{ij}$ were assumed
to be uniformly distributed in $[0,\,1]$.  To evaluate the
performance of our proposed decentralized stochastic gradient
descent algorithm, we set the stepsize of agent $i$ as
$\lambda_i^k=\frac{1-\varrho_i^k/k}{k}$ where $\varrho_i^k$ was
randomly chosen by agent $i$ from the interval $[0,\,1]$ in each
iteration. It can be verified that the stepsize satisfies the
summable conditions required in Theorem \ref{Theorem:Theorem_1}.
Every agent $i$ also chose $b_{ji}^k$ randomly in each iteration
under the sum-one condition. The evolution of the estimation error
averaged over 1,000 runs is illustrated by the solid blue line in
Fig. \ref{fig:comp_convex}. To compare the convergence performance
of our algorithm with the conventional decentralized stochastic
gradient descent algorithm, we also implemented the decentralized
stochastic gradient descent algorithm in \cite{lian2017can}, whose
average convergence trajectory over 1,000 runs is represented by the
black dotted line in Fig. \ref{fig:comp_convex}. {The
stepsizes of all agents were set to $\frac{1}{k}$}. Through
comparison, it can be seen that our introduced random parameters
$B^k$ and $\lambda_i^k$ do not slow down the convergence speed. In
fact, they significantly increase the optimization speed
 as well as optimization
accuracy compared with the
conventional decentralized SGD algorithm. 

\subsection{Non-convex case}
\begin{figure}
    \begin{center}
        \includegraphics[width=0.5 \textwidth]{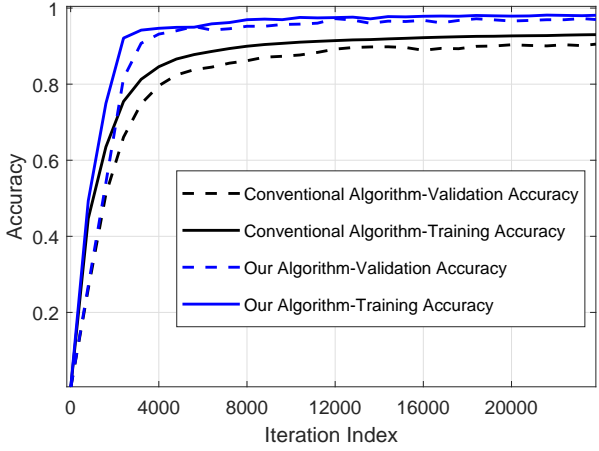}
    \end{center}
    \caption{Comparison of CNN training performance between our algorithm and the conventional decentralized SGD algorithm in \cite{lian2017can}.}
    \label{fig:nonconvex}
\end{figure}

We use the decentralized training of a convolutional neural network
(CNN)  to evaluate the performance of our proposed decentralized
stochastic gradient descent algorithm in non-convex optimization.
More specially, we   consider five agents interacting on a topology
depicted in Fig. \ref{fig:topology}. The agents collaboratively
train a CNN using the MNIST data set \cite{MNIST}, which is a large
benchmark database of handwritten digits widely used for training
and testing in the field of machine learning \cite{deng2012mnist}.
Each agent has a local copy of the CNN. The CNN has 2 convolutional
layers with 32 filters with each followed by a max pooling layer,
and then two more convolutional layers with 64 filters each followed
by another max pooling layer and a dense layer with 512 units. {  We used Sigmoid activation functions and hence the Lipschitz gradient assumption in Theorem \ref{theorem_non_convex} is satisfied. Note that ReLU activation functions will lead to non-smooth objective functions and hence non-Lipschitz gradients.} Each
agent has access to a portion of the MNIST data set, which was
further divided into two subsets for training and validation,
respectively.  We set the stepsize of agent $i$ as $\Lambda_i^k={\rm
diag}\{\frac{1-\varrho_{i1}^k/k}{k},\,\frac{1-\varrho_{i2}^k/k}{k},\cdots,\frac{1-\varrho_{id}^k/k}{k}\}$
where $\varrho_{ip}^k$ ($1\leq p\leq d$) were  statistically
independent random variables chosen by agent $i$ from the interval
$[0,\,1]$ in each iteration. (For the adopted CNN model, the
dimension of gradient $d$ is equal to $1,676,266$.) It can be
verified that the stepsizes satisfy the summable conditions required
in Theorem \ref{theorem_non_convex}. Every agent $i$ also chose
$b_{ji}^k$ randomly in each iteration under the sum-one condition.
The evolution of the training and validation accuracies averaged
over 1,000 runs are illustrated by the solid and dashed blue lines
in Fig. \ref{fig:nonconvex}. To compare the convergence performance
of our algorithm with the conventional decentralized stochastic
gradient descent algorithm, we also implemented the decentralized
stochastic gradient descent algorithm in \cite{lian2017can} to train
the same CNN {using stepsize $\frac{1}{k}$}, whose
average training and validation accuracies over 1,000 runs are
represented by the solid and dashed black lines in Fig.
\ref{fig:nonconvex}. It can be seen that the proposed algorithm has
a faster  converging speed as well as a slightly better
training/validation accuracy. 

\begin{figure}
    \begin{center}
        \includegraphics[width=0.5\textwidth]{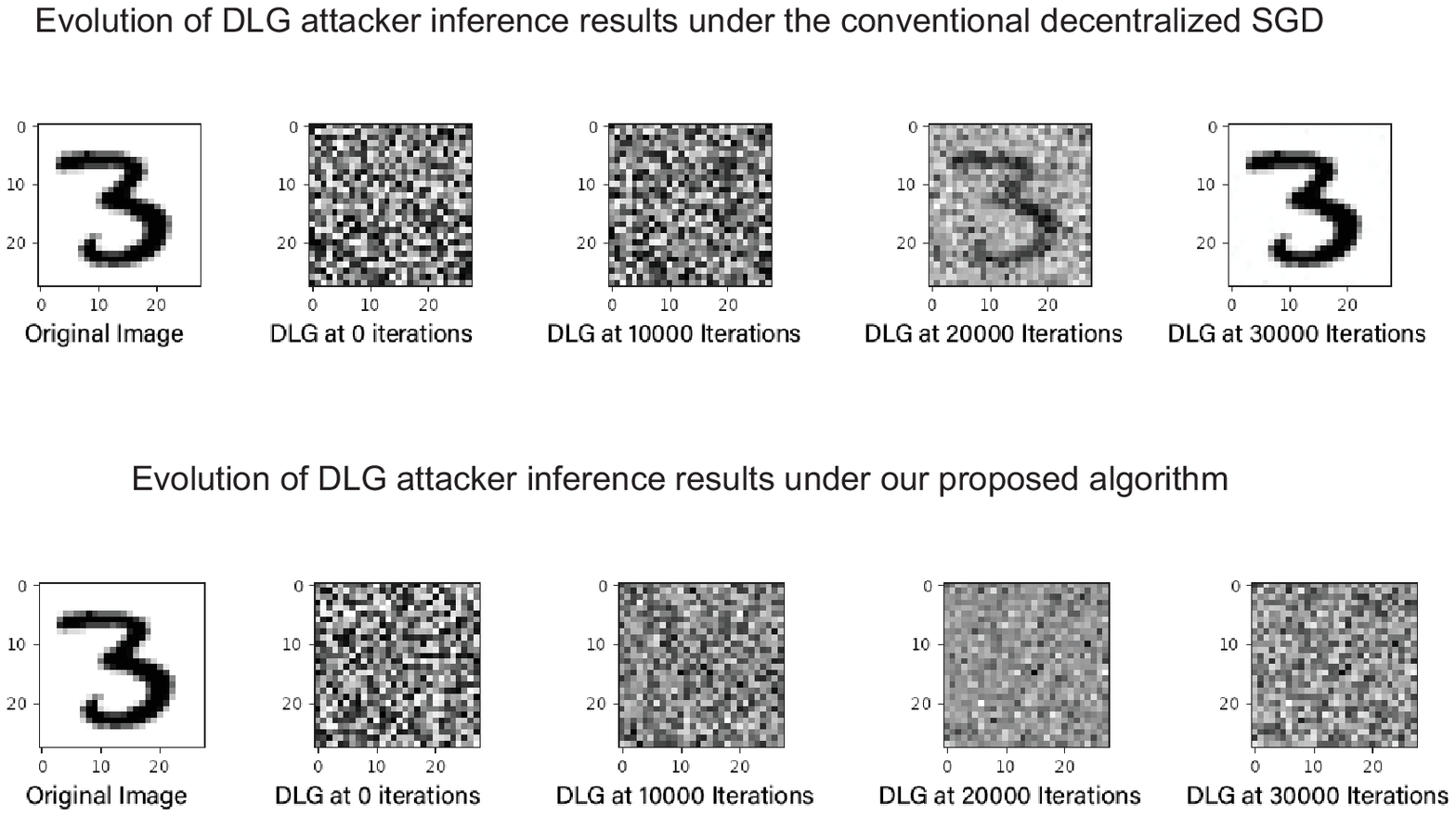}
    \end{center}
    \caption{Comparison of DLG attacher's inference results under the conventional decentralized SGD and our algorithm.}
    \label{fig:DLG_image}
\end{figure}

\begin{figure}
    \begin{center}
        \includegraphics[width=0.5\textwidth]{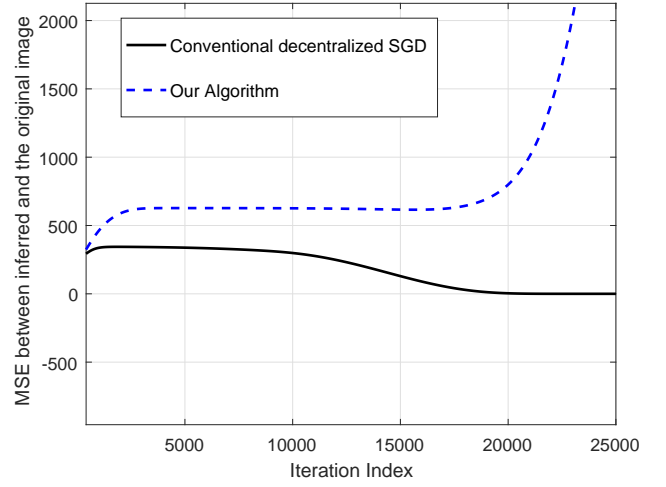}
    \end{center}
    \caption{Comparison of DLG attacher's inference errors under the conventional decentralized SGD and our algorithm.}
    \label{fig:DLG_MSE}
\end{figure}

To show that the proposed algorithm can indeed protect the privacy
of participating agents, we also implemented a privacy attacker
which tries to infer the raw image of participating agents using
received information. The attacker implements the DLG attack model
 proposed in \cite{zhu2019deep}, which is the most powerful
inference algorithm reported to date in terms of reconstructing
exact raw data from shared gradient/model updates. The attacker was
assumed to be able to eavesdrop all messages shared among the
agents. Fig. \ref{fig:DLG_image} shows that the attacker could
effectively recover the original training image from shared model
updates in the conventional stochastic gradient descent algorithm
\cite{lian2017can}. However, under the proposed algorithm, the
attacher failed to infer the original training image through
information shared in the network. This is also corroborated by the
attacker's inference performance measured by the mean-square error
(MSE) between the inference result  and the original image. More
specifically, as illustrated in Fig. \ref{fig:DLG_MSE}, the attacker
eventually inferred the raw image accurately as its estimation error
converged to zero. However, the proposed algorithm successfully
thwarted the attacker as attacker's estimation error was always
large.

{ Using the same setup, we  also compared our
privacy-preserving approach with a differential-privacy based
privacy approach that injects additive noise  to gradients directly
to protect privacy
\cite{yu2019differentially,bagdasaryan2019differential,mcmahan2018learning}.
More specifically,   we set $b_{ij}^k$ and $\Lambda_i^k$ in
(\ref{eq:proposed_algorithm}) as
$b_{ij}^k=\frac{1}{|\mathcal{N}_j|}$ (with $|\mathcal{N}_j|$
denoting the cardinality of $\mathcal{N}_j$) and
$\Lambda_i^k=\frac{1}{k}I_d$, respectively, and added $d$
dimensional zero-mean Gaussian noise $\xi_j$ directly to the
gradient $g_j^k$. We evaluated the training/validation accuracy and
DLG attacker's inference error  under different variances
$\sigma_{DP}$ of the differential-privacy noise $\xi_j$. The results
are summarized in Table I where every entry is the average over
1,000 runs.}
\begin{table*}
{\caption {Training/validation accuracy and DLG
attacker's inference errors under different levels of
differential-privacy noise}}
 \center
{
\begin{tabular}{  c|c|c|c|c|c|c  }
  &No Noise & $\sigma_{DP}=10^{-4}$ & $\sigma_{DP} =10^{-3}$&$\sigma_{DP}=10^{-2}$& $\sigma_{DP}=10^{-1}$&$\sigma_{DP}=10^{0}$ \\
  \hline
  Training Accuracy&0.9877&0.9823&0.7810& 0.0999 & 0.0995 &0.0995\\
  \hline
  Validation Accuracy &0.9879 & 0.9871& 0.7870& 0.0993 & 0.0992 & 0.0993\\
  \hline
  Final DLG Error & 3.030$\times 10^{-7}$& 0.1012 & 16.39 & 58.48 &340.3&1221
\end{tabular}}
\end{table*}

 {In the evaluation, we find that to avoid DLG
attackers from obtaining  images that are recognizable by human
eyes,  the DLG inference error should be at least 50. Therefore,
according to Table I, to combat DLG inference attacks,
 we have to use differential-privacy noise on the level of $10^{-2}$, which  significantly compromises both training and
validation accuracies. Given  that our privacy-preserving approach
can combat DLG attackers without sacrificing training or validation
accuracy (see Fig. \ref{fig:nonconvex} for  training and validation
accuracies and Fig. \ref{fig:DLG_MSE} for DLG attacker inference
error for our algorithm), the comparison clearly demonstrates the
potential of the proposed approach in enabling privacy protection.}

\section{Conclusions}
By judiciously manipulating inter-agent interaction  dynamics, this
paper has proposed an inherently privacy-preserving decentralized
stochastic gradient descent algorithm. The dynamics based privacy
design exploits the inherent robustness of decentralized stochastic
optimization to enable privacy while maintaining the accuracy of
optimization results, which is in distinct difference from
differential-privacy based privacy solutions that   trade
optimization accuracy for privacy. The inherently private
decentralized stochastic gradient descent algorithm is
encryption-free, and hence avoids incurring heavy extra
communication and computation overhead, an unavoidable problem with
existing encryption based privacy solutions for decentralized
stochastic optimization. Rigorous theoretical analysis has been
provided to confirm the accurate convergence  of all  decentralized
agents to a desired solution, both in the case with convex objective
functions and in the case with non-convex objective functions. An
information-theoretic  privacy analysis has also been proposed to
characterize the strength of privacy protection for each
participating agent against honest-but-curious and eavesdropping
adversaries. Both  simulation results for a convex decentralized
estimation problem and numerical experiments for decentralized
learning on a benchmark image dataset have confirmed the
effectiveness of the proposed algorithm. 

\section*{Acknowledgement}
The authors would like to thanks Ben Liggett for the help in numerical experiments. They would also like to thank the associate editor and anonymous reviewers, whose comments helped improve  the paper.

\section*{Appendix}

{ 

\begin{Lemma 2}\label{Lemma-polyak_2}(Lemma 2 in \cite{wang2022tailoring})
Let $\{v^k\}$,$\{\a^k\}$, and $\{p^k\}$ be random nonnegative scalar sequences, and
$\{q^k\}$ be a deterministic nonnegative scalar sequence satisfying
$\sum_{k=0}^\infty \a^k<\infty$ almost surely,
$\sum_{k=0}^\infty q^k=\infty$, $\sum_{k=0}^\infty p^k<\infty$ almost surely,
and the following inequality almost surely:
\[
\mathbb{E}\left[v^{k+1}|\mathcal{F}^k\right]\le(1+\a^k-q^k) v^k +p^k,\quad \forall k\geq 0
\]
where $\mathcal{F}^k=\{v^\ell,\a^\ell,p^\ell; 0\le \ell\le k\}$.
Then, we have $\sum_{k=0}^{\infty}q^k v^k<\infty$ and
$\lim_{k\to\infty} v^k=0$ almost surely.
\end{Lemma 2}

\begin{Lemma 3}\label{lem-opt}(Lemma 3 in \cite{wang2022tailoring})
Consider the problem $\min_{z \in \R^d} \phi(z)$,
where $\phi:\mathbb{R}^d\to\mathbb{R}$ is a
continuous function. Assume that
the optimal solution set $Z^*$ of the problem is nonempty.
Let $\{z^k\}$ be a random sequence such that for any optimal solution $z^*\in Z^*$,
\[
\begin{aligned}
&\mathbb{E}\left[\|z^{k+1}-z^*\|^2|\mathcal{F}^k\right]\\
&\le (1+\a^k)\|z^k - z^*\|^2  - \eta^k\left (\phi(z^k) - \phi(z^*)\right) +\b^k,\: \forall k\geq 0
\end{aligned}
\]
holds almost surely,
 where
$\mathcal{F}^k=\{z^\ell,\a^\ell,\b^\ell,\ \ell=0,1,\ldots,k\}$,
$\{\a^k\}$ and $\{\b^k\}$ are random nonnegative scalar sequences
satisfying $\sum_{k=0}^\infty \a^k<\infty$ and $\sum_{k=0}^\infty \b^k<\infty$ almost surely
while $\{\eta^k\}$ is a deterministic nonnegative scalar sequence with
$\sum_{k=0}^\infty \eta^k=\infty$.
Then, $\{z^k\}$ converges almost surely to some solution $z^*\in Z^*$.
\end{Lemma 3}
}

%
%
 \bibliographystyle{unsrt}

\bibliography{reference1}

\vspace{-1cm}
\begin{IEEEbiography}{Yongqiang Wang} (SM'13) was born in Shandong, China. He received the B.S. degree in electrical engineering and automation,
the B.S. degree in computer science and technology from Xi'an
Jiaotong University, Xi'an, Shaanxi, China, in 2004, and the M.Sc.
and Ph.D. degrees in control science and engineering from Tsinghua
University, Beijing, China, in 2009. From 2007-2008, he was with the
University of Duisburg-Essen, Germany, as a visiting student. He was
a Project Scientist at the University of California, Santa Barbara
before joining Clemson University, SC, USA, where he is currently an
Associate Professor. His current research interests include
decentralized control, optimization, and learning, with an emphasis
on privacy and security.   He currently serves as an associate
editor for {\it IEEE Transactions on Automatic Control} and {\it IEEE Transactions on Control of
Network Systems}.
\end{IEEEbiography}
\vspace{-1cm}

\begin{IEEEbiography}{H. Vincent Poor} (S72, M77, SM82, F87) received
the Ph.D. degree in EECS from Princeton University in 1977. From
1977 until 1990, he was on the faculty of the University of Illinois
at Urbana-Champaign. Since 1990 he has been on the faculty at
Princeton, where he is currently the Michael Henry Strater
University Professor. During 2006 to 2016, he served as the dean of
Princeton's School of Engineering and Applied Science. He has also
held visiting appointments at several other universities, including
most recently at Berkeley and Cambridge. His research interests are
in the areas of information theory, machine learning and network
science, and their applications in wireless networks, energy systems
and related fields. Among his publications in these areas is the
forthcoming book {\it Machine Learning and Wireless Communications}.
(Cambridge University Press).

Dr. Poor is a member of the National Academy of Engineering and the
National Academy of Sciences, and is a foreign member of the Chinese
Academy of Sciences, the Royal Society, and other national and
international academies. He received the IEEE Alexander Graham Bell
Medal in 2017.
\end{IEEEbiography}

\end{document}